\documentclass[nohyperref]{article}

\usepackage[protrusion=true,expansion=true,final]{microtype}
\usepackage{graphicx}
\usepackage{subfigure}
\usepackage{booktabs} 

\usepackage{hyperref}

\usepackage[accepted]{icml2023}

\usepackage{amsmath}
\usepackage{amssymb}
\usepackage{mathtools}
\usepackage{amsthm}

\usepackage[capitalize,noabbrev]{cleveref}

\theoremstyle{plain}

\theoremstyle{definition}

\theoremstyle{remark}

\usepackage[textsize=tiny]{todonotes}

\usepackage{booktabs}
\usepackage{xcolor}
\usepackage{algpseudocode}
\usepackage{tikz}
\newcommand{\ie}{\textit{i.e.}}
\newcommand{\Ours}{UPSCALE}
\newcommand{\OursTitle}{Unconstrained Channel Pruning}
\newcommand{\OursFull}{Unconstrained Channel Pruning Export}

\usepackage{pifont}

\usepackage[none]{hyphenat} 

\usetikzlibrary{decorations.pathreplacing,calc}
\newcommand{\tikzmark}[1]{\tikz[overlay,remember picture] \node (#1) {};}
\newcommand*{\AddNote}[4]{%
    \begin{tikzpicture}[overlay, remember picture]
        \draw [decoration={brace,amplitude=0.5em,raise=1ex},decorate,thick,gray]
            ($(#3)!(#2.south)!($(#3)-(0,1)$)$) --
            ($(#3)!(#1.north)!($(#3)-(0,1)$)$)
                node [align=center, text width=3cm, pos=0.5, anchor=east, font=\it, rotate=90, yshift=4ex, xshift=11ex] {#4};
    \end{tikzpicture}
}%

\icmltitlerunning{\OursTitle}

\begin{document}

\twocolumn[
\icmltitle{\Ours: \OursTitle}



\icmlsetsymbol{equal}{*}

\begin{icmlauthorlist}
\icmlauthor{Alvin Wan}{}
\icmlauthor{Hanxiang Hao}{}
\icmlauthor{Kaushik Patnaik}{}
\icmlauthor{Yueyang Xu}{}
\icmlauthor{Omer Hadad}{}
\icmlauthor{David G\"uera}{}
\icmlauthor{Zhile Ren}{}
\icmlauthor{Qi Shan}{apple}
\end{icmlauthorlist}

\icmlaffiliation{apple}{Apple, Cupertino, USA}

\icmlcorrespondingauthor{Alvin Wan}{alvinwan@apple.com}
\icmlcorrespondingauthor{Qi Shan}{qshan@apple.com}

\icmlkeywords{Machine Learning, ICML}

\vskip 0.3in
]



\printAffiliationsAndNotice{}  

\begin{abstract}
As neural networks grow in size and complexity, inference speeds decline. To combat this, one of the most effective compression techniques -- channel pruning -- removes channels from weights. However, for multi-branch segments of a model, channel removal can introduce inference-time memory copies. In turn, these copies increase inference latency -- so much so that the \textit{pruned} model can be slower than the \textit{unpruned} model. As a workaround, pruners conventionally constrain certain channels to be pruned together. This fully eliminates memory copies but, as we show, significantly impairs accuracy. We now have a dilemma: Remove constraints but increase latency, or add constraints and impair accuracy. In response, our insight is to reorder channels at export time, (1) reducing latency by reducing memory copies and (2) improving accuracy by removing constraints. Using this insight, we design a generic algorithm \Ours\footnote{\url{https://github.com/apple/ml-upscale}}~to prune models with \textit{any} pruning pattern. By removing constraints from existing pruners, we improve ImageNet accuracy for post-training pruned models by 2.1 points on average -- benefiting DenseNet (+16.9), EfficientNetV2 (+7.9), and ResNet (+6.2). Furthermore, by reordering channels, \Ours~improves inference speeds by up to 2$\times$ over a baseline export.
\end{abstract}

\begin{figure}[!ht]
    \centering
    \includegraphics[width=0.47\textwidth]{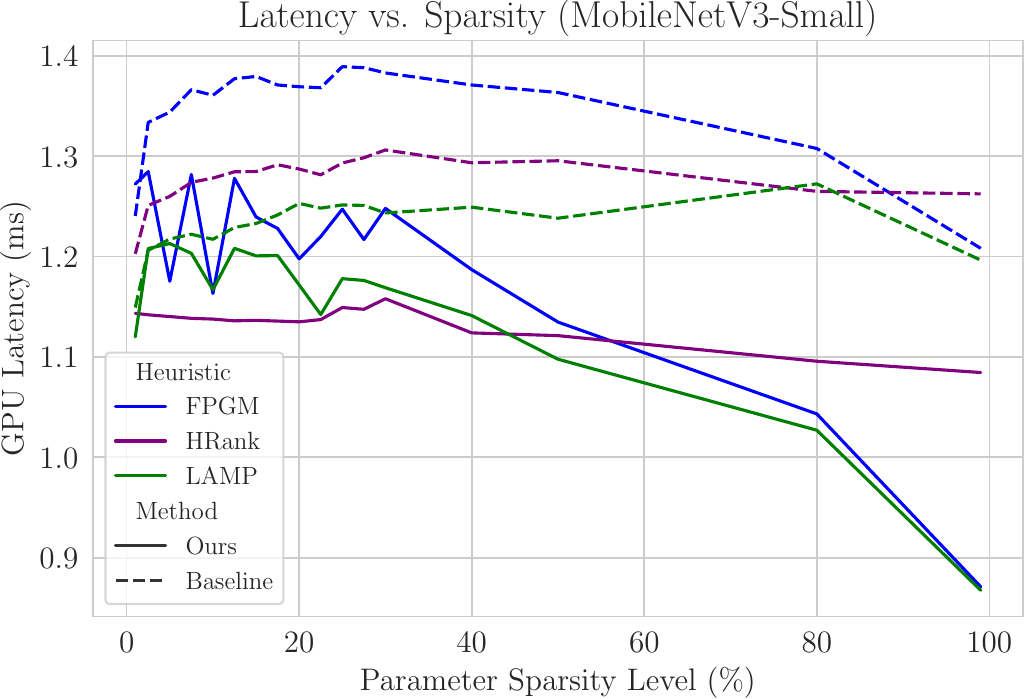}
    \caption{\textbf{Our export achieves lower latency} (\Ours, solid) than the baseline export (dotted) for an unconstrained pruned model. Notice the baseline export (dotted) is unable to achieve a latency lower than the \textit{unpruned} model's latency of $\sim$1.1ms, even with over 95\% sparsity levels. By contrast, our export (solid) reduces latency more appropriately. The above plot features MobileNetV3-Small pruned at various sparsity levels, using existing pruning heuristics (FPGM, HRank, LAMP) but modified to run without channel constraints.}
\end{figure}

\begin{figure*}
    \centering
    \includegraphics[width=\textwidth]{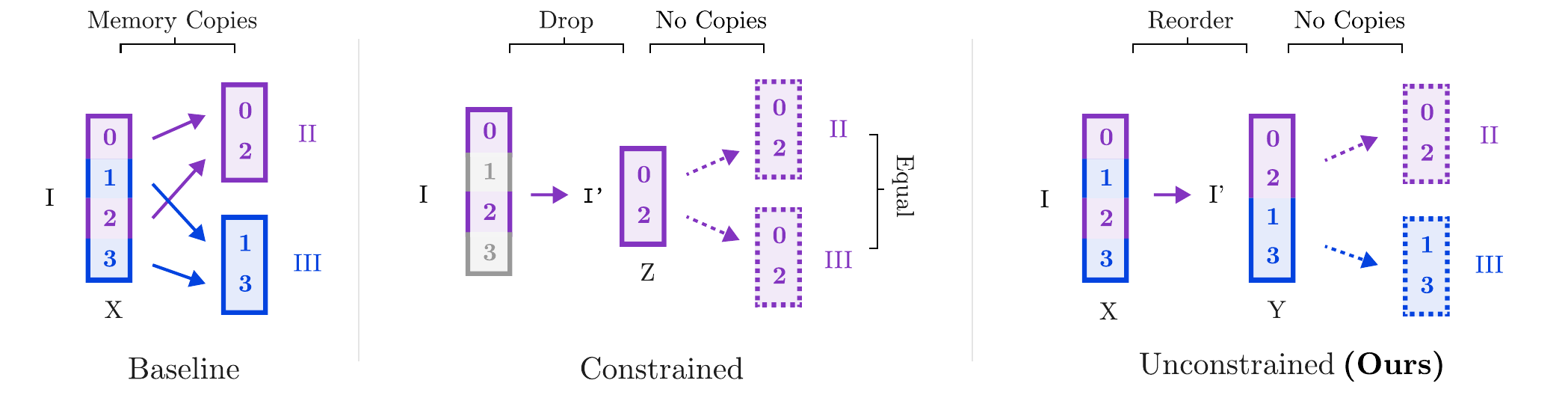}
    \vspace{-15pt}
    \caption{\textbf{Unconstrained pruning patterns are non-trivial to export.} \textit{Left}: Different convolutions (II, III) prune different channels in a shared tensor $X$. As a result, channels from $X$ are copied during \textit{inference} time to assemble inputs for II (purple), III (blue). Unfortunately, these memory copies incur significant latency costs. \textit{Middle}: To sidestep this, previous methods constrain all convolutions (II, III) to prune the same channels. Since all convolutions prune the same channels, the common, pruned filters can be removed before inference, converting convolution I into I'. However, these constraints limit the pruned model's accuracy. \textit{Right}: To eliminate memory copies without constraints, we reorder output channels in convolution I to generate I'. This way, inputs for downstream convolutions (II, III) are contiguous in memory, in $Y$. Our \Ours~algorithm computes this reordering.}
    \label{fig:teaser}
\end{figure*}

\begin{figure}[!ht]
    \centering
    \includegraphics[width=0.47\textwidth]{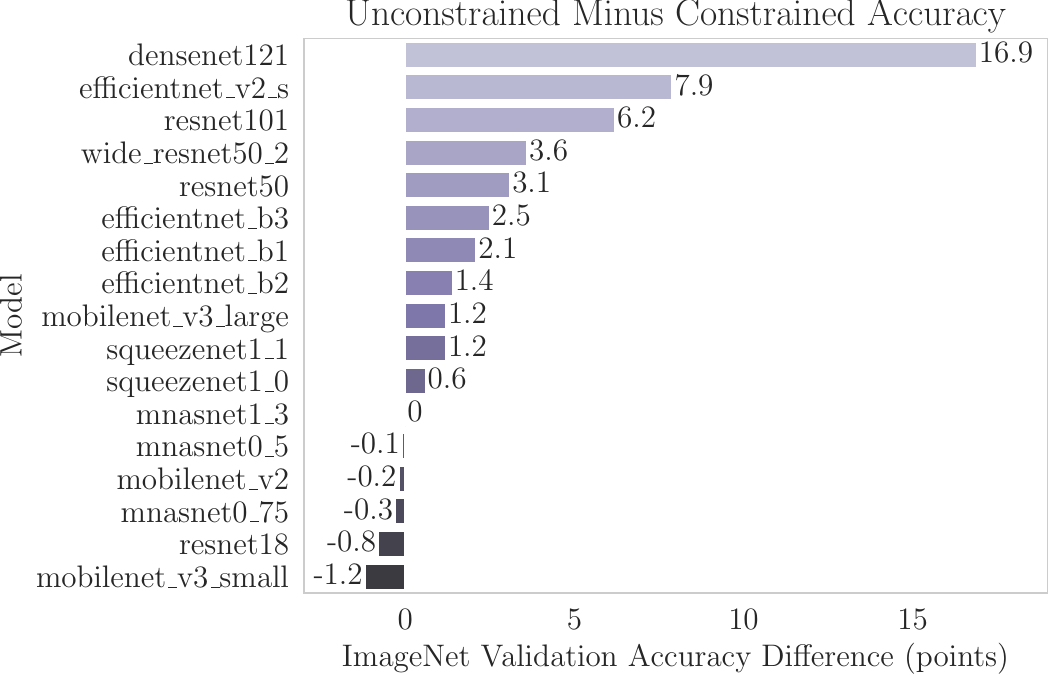}
    \caption{\textbf{Removing pruning constraints improves accuracy}. Above is the accuracy of unconstrained minus constrained pruned models, averaged across sparsity levels and heuristics. Positive values mean that unconstrained pruning outperforms constrained pruning. For example, on DenseNet121, unconstrained outperforms constrained pruning by an average of 16.9 points. This is generally true as the model grows larger; we hypothesize this is due to larger models with more channels having more constraints. For example, ResNet18 (-0.8) favors constrained pruning, but its larger variants ResNet50 (+3.1) and ResNet101 (+6.2) both favor unconstrained pruning. The same pattern repeats for MobileNetV3.}
    \label{fig:unconstrained-larger-models}
\end{figure}

\vspace{-10pt}

\section{Introduction}
\label{sec:intro}

Deep neural networks are intrinsic to an increasing number of real-world applications, but neural network architectures are simultaneously growing in size and complexity, year over year. This is problematic for a growing number of uses cases with inference-time resource constraints, making model compression techniques a necessity.

One of the most effective families of model compression methods is channel pruning, which removes entire channels of convolutional or dense weights to reduce inference-time latency. Despite its effectiveness, the process of channel removal itself -- which we refer to as the export step -- is non-trivial and is often overlooked. This leads to two disagreeable options for channel pruning: (1) Add constraints, per convention, that impede accuracy, or (2) remove constraints and unsustainably increase inference-time latency.

First, \textbf{without constraints, exporting is non-trivial} specifically for multi-branch segments of network. In these segments, layers from different branches may all use the same input feature map and simultaneously prune different channels. This causes incompatible dimensions and more importantly, during inference time, the network must then perform memory copies to ensure that tensors are contiguous in memory. These memory copies are latency hungry -- so much so that pruned models can be even slower than unpruned models (Figure \ref{fig:teaser}).

Second, to address this, \textbf{early works established convention by adding constraints} \cite{l1,l2} -- specifically, constrain layers in all branches of a segment to prune the same channels. These constraints simplify export, and as a result, modern pruning works focus on \textit{which} channels to prune rather than \textit{how} to remove them during export. Yet, despite significant progress in structured pruning, these constraints impair accuracy \cite{liu2018rethinking}. Intuitively, constraints restrict the set of possible pruning patterns, limiting pruning's effectiveness (Figure \ref{fig:unconstrained-larger-models}).

To tackle this problem, our insight is two-fold: (1) reordering channels can keep subsets of the tensor contiguous and (2) contiguous slices of a tensor are ``free'' to use, without memory copies. In short, this allows us to abandon conventional pruning constraints to obtain higher accuracy. Furthermore, by eliminating memory copies, this allows accuracy gains to come at a reduced latency penalty.

These insights yield a general-purpose export algorithm \OursFull~(\Ours)~that can prune models with \textit{any} pruning pattern. This enables a broader class of possible pruning algorithms by making latency for unconstrained pruned models much more palatable -- in some cases, almost fully eliminating extraneous memory copies. This work yields the following three contributions:
\begin{enumerate}
    \item \textbf{Comparison of unconstrained and constrained pruning} accuracy, on a variety of pruning heuristics, architectures, and sparsity levels. Unconstrained pruning improves post-training pruned accuracy by an average of 2.1 points on ImageNet.
    \item \textbf{A generic export utility for unconstrained pruning} dubbed \Ours, which produces models with inference times over $2\times$ faster than baseline export. This drop-in utility enables future researchers to abstract away export concerns. To the best of our knowledge, we are the first to develop an export utility that can support unconstrained pruning patterns.
    \item \textbf{Graph formulation of memory copy maximization}, an abstraction that may open opportunities for further latency optimizations.
\end{enumerate}

\begin{figure*}
    \centering
    \includegraphics[width=\textwidth]{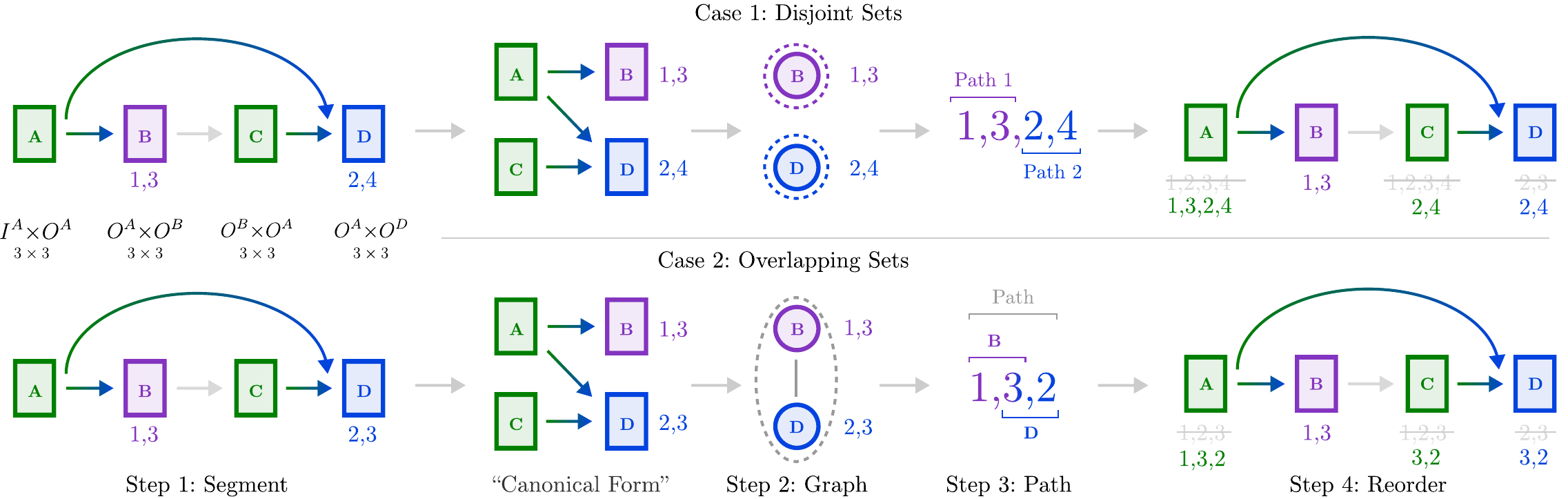}
    \caption{\textbf{Pipeline to export a model pruned without constraints:} \textit{Top:} Consider a typical residual block on the left. Underneath, we list the in and out channels, as well as the kernel size. In \textit{Step 1}, extract a \textit{segment}: This includes the $A \rightarrow B, A \rightarrow D, C \rightarrow D$ edges. Convolution B retains channels 1, 3; D retains 2, 4. Notice B and D retain disjoint sets of channels. Redraw in ``canonical'' form, with producers on the left and consumers on the right. This helps visualize constraints. In \textit{Step 2}, construct a graph. Since B and D don't retain any shared channels, they don't share an edge. There are now two paths, each just a single node (dotted circles). In \textit{Step 3}, order the channels. Simply place the channels for each path one after another. In \textit{Step 4}, use this channel ordering to reorder channels for all convolutions. \textit{Bottom}: In \textit{Step 1}, extract a segment. This time, D retains 2,3. Notice B and D share a retained channel (3). In \textit{Step 2}, draw two nodes for the two consumers B and C. Since B and D both retain 3, they share an edge. Find a maximum reward acyclic path, which includes the only two nodes (gray, dotted line). In \textit{Step 3}, list channels unique to B (1), those shared by B and D (3), then those unique to D (2). In \textit{Step 4}, reorder channels.}
    \label{fig:unified}
\end{figure*}

\section{Related Work}

\textbf{Structured vs. Unstructured Pruning}. Existing pruning work can be divided into unstructured methods \cite{han2016deep, molchanov2017variational,janowsky1989pruning,frantar2022spdy,yu2022combinatorial,he2022sparse} and structured methods \cite{anwar2017structured,He_2017_ICCV,sze2017efficient,molchanov2016pruning,zhu2017prune,huang2021training}. Unstructured methods threshold and zero individual weight values, whereas structured methods impose structure on sparsity patterns at various levels of granularity -- from pruning in blocks \cite{narang2017block} to removing entire channels \cite{lamp,hrank} -- the latter of which is called channel pruning \cite{l2}. Channel pruning is a unique pruning strategy, as its benefits are not dependent on custom sparse matrix primitives \cite{bulucc2012parallel,azad2017work,baskaran2009optimizing,vuduc2005oski}. Instead, channel pruning's benefits stem from removing convolutional weight channels at export time, reducing resource consumption at inference time.

\textbf{Channel Pruning Strategies} In addition to its unique benefit, channel pruning introduces a unique challenge: removing channels is difficult for complex topologies, especially when multiple branches in a network use or produce a shared tensor. In response, prior work \cite{he2017channel,l2,l1} imposes constraints on pruning masks, restricting the set of possible pruning patterns in complex portions of a network. Most work in structured pruning focuses on other aspects of pruning, other than exporting -- including heuristics for importance, whether local to a segment of the network \cite{han2015learning} or global, across layers and channels \cite{lee2018snip,frankle2018lottery}; the amount to sparsify at once \cite{liu2017learning,gale2019state}; and techniques for learning masks, whether using additional networks or layers \cite{huang2018learning,yamamoto2018pcas,he2018amc} or auxiliary variables \cite{guo2016dynamic,savarese2020winning,courbariaux2015binaryconnect,louizos2017learning,srinivas2017training,xiao2019autoprune,bengio2013estimating}.

Notably, these works overlook the export challenge for two reasons: Most works (1) apply zero-one masks during training to mimic pruning~\cite{fpgm,he2017channel,yu2022topology}, without ultimately removing channels; this means the challenge of dropping channels from multi-branch segments is never surfaced. Most works furthermore (2) report FLOPs as a proxy for latency \cite{fpgm,reed1993pruning,blalock2020state,Li_2022_CVPR,miao2021learning,wang2022recent,li2022pruning,shang2022neural}; this means that the latency impact of memory copies would have never been realized, had these works removed constraints. The combination of both practices thus hides export-related challenges. To remedy this, we study export directly to allow a broader class of pruning algorithms, showing that existing pruning heuristics can attain higher accuracy by simply removing these constraints.

\begin{algorithm}[!ht]
    \caption{\Ours}
    \footnotesize
    \begin{algorithmic}[1]
\Procedure{\Ours}{model}
    \State $L \gets$ model.layers
    \While{$L$}
        \State \tikzmark{left0}\tikzmark{top0} producers $\gets$ $\{L\textrm{.pop()}\} $\Comment{\textit{Sec \ref{sec:dividing-into-segments}}}
        \State consumers $\gets$ \{\}
        \State $N \gets$ 0
        \While{\textsc{Len}(producers) + \textsc{Len}(consumers) $> N$}
            \State N $\gets$ \textsc{Len}(producers) + \textsc{Len}(consumers)
            \State consumers $\gets$ \textsc{Consumers}(producers)
            \State producers $\gets$ \textsc{Producers}(consumers)
        \EndWhile
        \tikzmark{bottom0}

        \State \tikzmark{left1}\tikzmark{top1} $c \gets$ model.channels
        \State $G \gets$ consumers \Comment{\textit{Sec \ref{sec:reordering-as-a-graph-problem}}}
        \For{$n \in G$}
            \State $R_n[n] \gets \textsc{len}(c[n])$
        \EndFor
        \For{$m, n \in G$}
            \State $R_e[m,n] \gets -\textsc{len}(c[m] \cap c[n])$
        \EndFor
        \State $A \gets R_e > 0$\Comment{\textit{edge if $> 0$ shared channels}}
        \State $\Pi_i \gets$ \textsc{Reduce}(producers)\Comment{\textit{Sec \ref{sec:edge-cases}}}
        \tikzmark{bottom1}

        \State \tikzmark{left2}\tikzmark{top2} o $\gets$ []
        \While{$G$}\Comment{\textit{Sec. \ref{sec:solving-graph-problem}}}
            \State $p \gets \textsc{Mrap}(G)$\Comment{\textit{Eqn. \ref{eqn:mrap}}}
            \For{$n_i$ \textbf{in}~$p$}\Comment{\textit{Sec. \ref{sec:ordering}}}
                \State $\tilde{n_i} \gets n_i - n_{i-1} - n_{i+1}$\Comment{\textit{Eqn. \ref{eqn:channels-unique-to-node}}}
                \State o.extend($\tilde{n_i} \cup (n_i \cap n_{i+1})$) \Comment{\textit{Eqn. \ref{eqn:channel-ordering}}}
            \EndFor
            \State $G \gets G - p$
        \EndWhile
        \tikzmark{bottom2}

        \State \tikzmark{left3}\tikzmark{top3}$W \gets$ model.weights
        \State $\Pi \gets \pi(o)$\Comment{\textit{to permutation matrix}}
        \For{$p_i$ \textbf{in} producers} \Comment{\textit{Sec \ref{sec:reordering-weights}}}
            \State $W[p_i] \gets W[p_i] \Pi_i \Pi$
        \EndFor
        \For{$c_i$ \textbf{in} consumers}
            \State $W[c_i] \gets W[c_i] \Pi$
        \EndFor
        \tikzmark{bottom3}
    \EndWhile
\EndProcedure
    \end{algorithmic}
    \AddNote{top0}{bottom0}{left0}{1. segment net}
    \AddNote{top1}{bottom1}{left1}{2. define graph}
    \AddNote{top2}{bottom2}{left2}{3. compute order}
    \AddNote{top3}{bottom3}{left3}{4. reorder weights}
    \label{alg:upscale}
\end{algorithm}

\section{Method}
\label{sec:method}

Previous work focuses on pruning \textit{strategies}, which mimics pruning patterns by ``zero'ing'', channels during training. Our work is orthogonal to this line of work, instead focusing on how to \textit{export} unconstrained pruning patterns. 

Both the input and output channels in a kernel can be pruned; pruning the former is called ``input pruning'' (\ie, place the pruning mask before the convolution layer), and pruning the latter is called ``output pruning'' (\ie, place the pruning mask after the convolution layer). Without loss of generality, we will focus on unconstrained \textit{input} pruning. To see an explicit description of unconstrained output pruning, see Appendix \ref{sec:unconstrained-output-pruning}.
We also empirically observe that input pruning generally outperforms output pruning (Figures A.\ref{fig:input_vs_output_accuracy_1}, A.\ref{fig:input_vs_output_accuracy_2}), motivating our focus on input pruning.

Note that throughout this methods section, we explicitly specify ``convolutions'' for ease of understanding. However, our algorithm generalizes to any operation where the number of input channels is independent from the number of output channels, such as dense layers.

\subsection{Step 1 - Reduce to a Segment}
\label{sec:dividing-into-segments}

Our algorithm divides the network architecture into \textit{segments} (Figure \ref{fig:unified}, Step 1) -- informally, a set of layers that can be pruned independently of the rest of the network. A segment includes (a) convolutions that produce outputs, dubbed \textit{producers} and (b) convolutions that consume those tensors, dubbed \textit{consumers}.

To identify segments, start from an arbitrary producer. Find all consumers for that producer, then find all producers for those consumers. Repeat iteratively until both sets converge. See Algorithm \ref{alg:upscale} for a more exact formulation.

For example, in Figure \ref{fig:unified} (Step 1), start from producer $A$. Find $A$'s consumers: $\{B, D\}$. Find their producers: $\{A, C\}$. Find their consumers: $\{B, D\}$. Notice our set of consumers has converged, so our segment is complete. This segment can now be pruned independently of the rest of the network, reducing the problem of pruning a network to pruning a single segment. See Figure A.\ref{fig:simple-vs-complex-segments} for more examples.

\subsection{Why Memory Copies Occur}
\label{sec:why-memory-copies-occur}

In a nutshell, memory copies occur when consumers prune different channels. To handle this, prior methods simply constrain all consumers to prune the same channels, as in Figure \ref{fig:teaser}, ``Constrained''. Since both consumers are constrained to prune the same channels, this is trivial to export: Simply drop the producer's output channels that all consumers prune.

However, consider now the unconstrained case in Figure \ref{fig:teaser}, ``Inefficient''. Say consumer II prunes channels that consumer III does not. At export time, we cannot drop channels that II prunes, since III still needs it. Instead, during \textit{inference} time, we subselect the producer's output -- copy every unpruned channel that II needs into a new tensor -- to then pass to II. This method is the baseline approach to exporting unconstrained pruned models.
Unfortunately, the memory copies in the baseline approach incur significant latency costs at \textit{inference} time, so much so that this pruned model incurs \textit{higher} latency than the original, unpruned model (Figure \ref{fig:reordering_latency}).

Our key insight is that contiguous slices are ``free''. In other words, each consumer takes a slice of the input tensor, instead of copying channels into a new tensor. We therefore design an algorithm that reorders producer output channels and consumer input channels (Figure \ref{fig:teaser}, ``Ours'', Sec \ref{sec:reordering-weights}). We now discuss how to compute this reordering.

\begin{figure}
    \includegraphics[width=.47\textwidth]{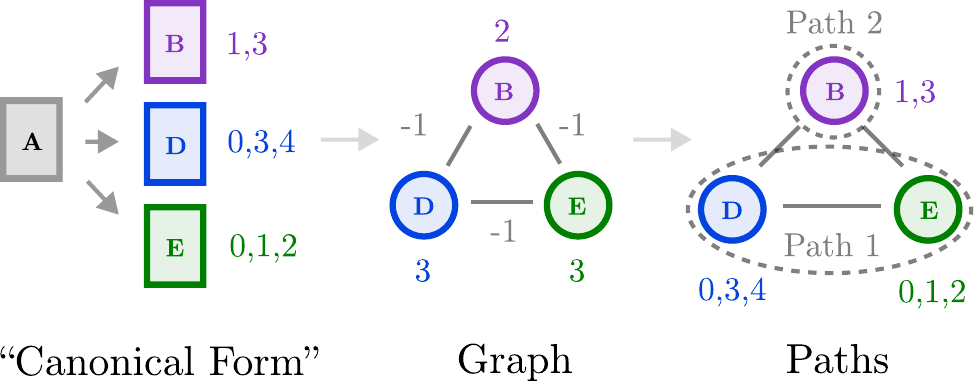}
    \caption{\textbf{Dividing a graph into maximum reward acyclic paths}. This is a more detailed version of Fig \ref{fig:unified}, where we show a slightly more involved example. \textit{Left}: We have a segment in canonical form, with 1 producer and 3 consumers. \textit{Center}: We build the graph, with one node per consumer. Notice every consumer shares a channel with every other consumer. This means the graph is fully-connected, creating a cycle. B has a reward of 2 since B retains 2 channels. D and E both retain 3 channels so have a reward of 3. Every pair shares exactly 1 retained channel, so every edge has a reward of -1. \textit{Right}: We find the maximum reward acyclic path. Since the graph is fully-connected, any pair of nodes is a viable path: BD, BE, or DE. Of the 3, DE has the highest reward (3 + 3 - 1), so DE is ``Path 1''. B is then ``Path 2''.}
    \label{fig:mrap}
\end{figure}

\subsection{Step 2 - Formulate as a Graph}
\label{sec:reordering-as-a-graph-problem}

To simplify explanations, moving forward, we will refer to the channels that a consumer \textit{retains}, rather than the channels that it prunes.

To formalize our objective, we formulate our constraints as an undirected graph (Figure \ref{fig:unified}, Step 2): Each node in our graph represents the set of channels that a consumer retains. Two nodes share an undirected edge if both nodes share one more more retained channels. Each node's reward is the number of retained channels, and each edge's reward is \textit{negative} of the number of shared retained channels. We will discuss why shortly. With this graph, we can formalize our memory-copy maximization objective.

\textbf{Our goal is to find a path} -- or in other words, a sequence of nodes: Since each node is a consumer, a sequence of nodes represents a sequence of consumers. In turn, every sequence of consumers admits a zero-copy order of channels. For example, say we have two consumers with node sequence $A, B$. The zero-copy channel order can be obtained by: ordering channels retained only by $A$, then channels shared by $A$ and $B$, then channels retained only by $B$. This ordering ensures zero memory copies, as both $A$ and $B$'s inputs are already contiguous in memory. We can continue indefinitely for any number of consumers, so long as the consumers are ordered in a sequence. In sum, a path can be translated into a zero-copy channel ordering. For more examples, see Appendix \ref{sec:reordering-examples}.

\textbf{Our goal is to find an \textit{acyclic} path}: Say the first and last nodes in our path share channels. Now, there's a dilemma: Place the shared channels either at the beginning or the end of the channel ordering. Either way, at least one consumer will have its input spread non-contiguously in memory.  More generally, non-adjacent consumers in the sequence cannot share channels. In other words, no two non-adjacent nodes can share an edge -- or more succinctly, the path must be acyclic; otherwise, we will need to introduce additional memory copies for the shared channels during inference.

\textbf{Our goal is to find the \textit{maximum reward} acyclic path}: As mentioned earlier, all channels in a path require zero copies, so to minimize memory copies, maximize the number of channels included in the path. Previously, we defined node and edge reward so that path reward is equal to the number of channels included in a path. As a result, in turn, maximizing included channels maximizes path reward. 

Our final, formal objective is to find the maximum reward acyclic path. For end-to-end examples of how this minimizes memory copies, see Appendix \ref{sec:graph-algorithm-examples} or Figure A.\ref{fig:graphs}.

\begin{figure*}[!ht]
    \centering
    \includegraphics[width=\textwidth]{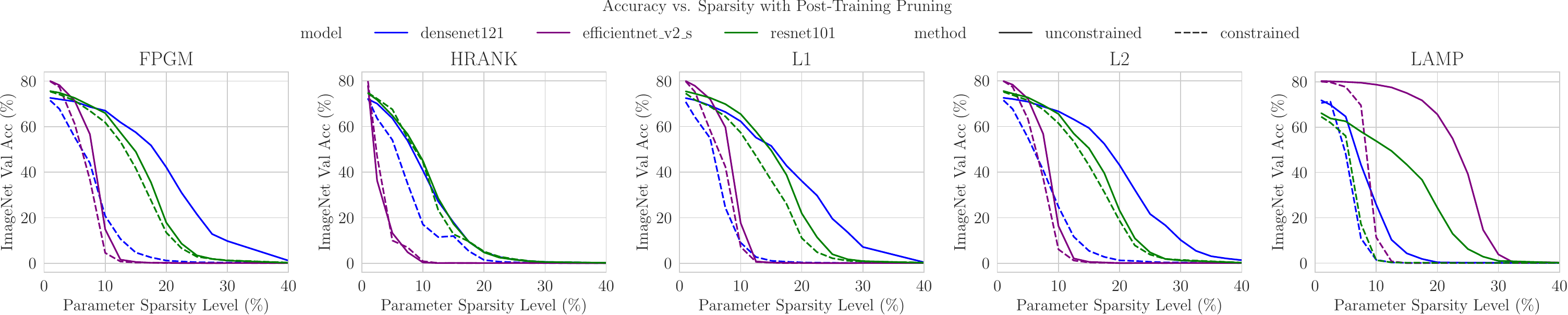}
    \caption{\textbf{Unconstrained Outperforms Constrained Pruning Accuracy} for larger models and models with complex topologies. The above models are pruned, then evaluated on ImageNet's validation set across various sparsity levels. Unconstrained pruning (solid) achieves higher accuracy than constrained pruning (dashed), with benefits varying depending on the pruning strategy. Per figures A.\ref{fig:input_vs_output_accuracy_1}, A.\ref{fig:input_vs_output_accuracy_2} input pruning outperforms output pruning in our experiments, so we narrow our focus to input pruning in the main text. Pruning heuristics include L1 \cite{l1}, L2 \cite{l2}, LAMP \cite{lamp}, FPGM \cite{fpgm}, and HRank \cite{hrank}. Architectures include DenseNet \cite{densenet}, EfficientNetV2 \cite{efficientnetv2}, ResNet \cite{resnet}. See all 80+ plots across architectures and heuristics in figures A.\ref{fig:unconstrained_vs_constrained_accuracy_1}, A.\ref{fig:unconstrained_vs_constrained_accuracy_2}.} 
    \label{fig:unconstrained-vs-constrained-accuracy-main}
\end{figure*}

\subsection{How to Find Maximum Reward Acyclic Path}
\label{sec:solving-graph-problem}

We start by computing the reward of the maximum reward path, which we abbreviate as ``mrp''. To compute this reward $\textsc{Mrp}(G)$ for a graph $G$, iterate over all nodes $s \in G$, and compute reward $f(s)$ of the mrp \textit{starting from node $s$}.
\begin{equation}
\label{eqn:mrp}
\textsc{Mrp}(G) = \textsc{max}(\{f(s) : s \in G\})
\end{equation}

Our goal is now to define $f(s)$. To do so, consider all of node $s$'s neighbors: $n \in A[s]$, for adjacency matrix $A$. Intuitively, the reward $f(s)$ of the mrp is the maximum (a) reward of all maximum reward paths starting from its neighboring nodes $f(n) : n \in A[s]$, along with (b) the reward of traveling to its neighboring nodes $R_e[s, n]$, for edge reward matrix $R_e$. Recall that $R_e$ is the number of shared channels. This is summarized in the following recurrence relation
\begin{equation}
f(s) = \textsc{max}(\{f(n) + R_e[s, n]: n \in A[s]\}) + R_n[s]
\end{equation}

for node reward matrix $R_n$. Recall $R_n$ is the number of channels that a consumer retains. This finds the maximum reward path, but it does not handle the acyclic constraint. Note that the mrp path $p$ is the node sequence obtained by optimizing the objective (i.e., Eqn \ref{eqn:mrp}) via dynamic programming.

\textbf{Handling the acyclic constraint}: We abbreviate maximum reward acyclic path as ``mrap''. To handle the acyclic constraint, we define a set of nodes $I$, which includes all invalid nodes. Invalid nodes include (a) already-traversed nodes included in the current path, as well as (b) nodes neighboring the path -- \ie, nodes sharing an edge with a node in the path. Our path is guaranteed to be acyclic if we never traverse nodes neighboring our path. We redefine our subproblem to be $f(s, I)$, the reward of the mrap starting from source node $s$ and excluding nodes in $I$.

We can then amend our original objective to be the following. The initial invalid set $I$ is the only node in the path, $\{s\}$.

\begin{equation}
\label{eqn:mrap}
\textsc{Mrap}(G) = \textsc{max}(\{f(s, \underbrace{\{s\}}_{I}) : s \in G\})
\end{equation}

We can then amend the recurrence relation to be the following. For convenience, we factor out the definition for $g(s, n, I)$, which is the reward of traveling only from $s$ to $n$ while avoiding invalid set $I$.

\begin{align}
\label{eqn:dp-invalids}
f(s, I) &= \textsc{max}(\{g(s, I, n): n \in \overbrace{A[s] - I}^{\textrm{ignoring }I}\}) + R_n[s] \notag \\
g(s, I, n) &= f(n, \underbrace{I \cup \{n\} \cup A[s]}_{\textrm{expanding }I}) + R_e[s, n] 
\end{align}

Less formally, Eqn \ref{eqn:dp-invalids} excludes all invalid nodes when considering neighbors to traverse to. When considering each neighboring node, we expand the invalid set $I$ by adding the neighboring node itself $\{n\}$, as well the source node's neighbors $A[s]$.

\begin{figure*}
    \centering
    \includegraphics[width=\textwidth]{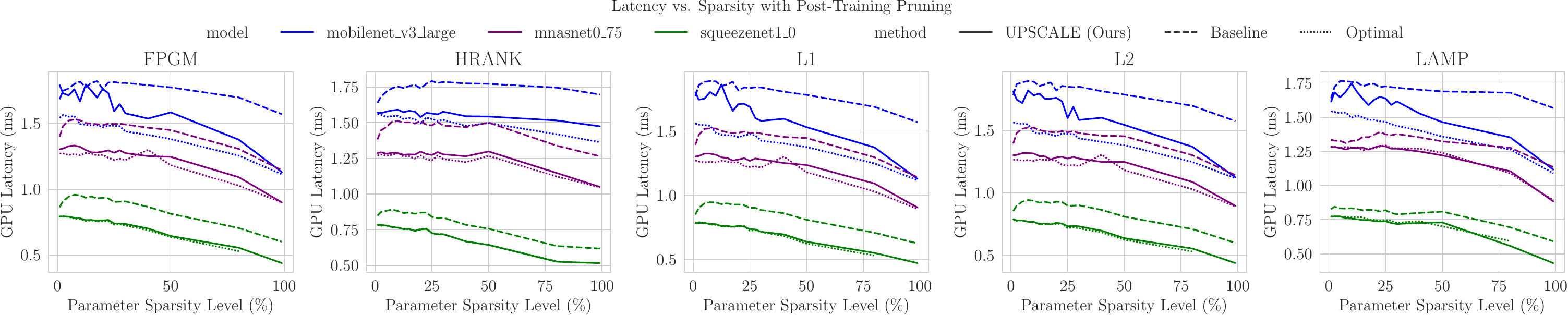}
    \caption{\textbf{Necessity of Reordering for Latency Improvement}: We observe that baseline unconstrained pruning export (dashed) can actually be detrimental to total latency, even increasing latency beyond that of the \textit{unpruned} model. By contrast, using \Ours~(solid), the pruned model sees more appropriate latency reductions. To visualize the upper bound on \Ours's latency savings, we include latency for a theoretical, zero-memory-copy export (dotted). \Ours~vastly outperforms the baseline export and for mobile models such as SqueezeNet (green), \Ours's latency (solid green) even approaches the optimal latency reduction (dotted green). Architectures include MobileNetV3 \cite{mobilenetv3}, MnasNet \cite{mnasnet}, SqueezeNet \cite{squeezenet}. See all plots in Figure A.\ref{fig:unconstrained_vs_constrained_latency}.}
    \label{fig:reordering_latency}
\end{figure*}
 
\subsection{Step 3 - Compute Channel Order From Graph}
\label{sec:ordering}

Note that the maximum reward acyclic path $p$ can be obtained by solving the optimization (i.e., Eqn \ref{eqn:mrap}) via dynamic programming, which is denoted as $p \gets \textsc{Mrap}(G)$. Furthermore, the mrap path $p$ may not include all nodes -- for example, if all nodes lie in a ring. As a result, continue to find paths on all remaining nodes $\textsc{Mrap}(G - p)$ until there are no more nodes remaining.

The paths we find in the previous step are then translated into a final channel ordering. As in Section~\ref{sec:reordering-as-a-graph-problem} and Figure \ref{fig:unified} (Case 2), say we have two consumers, $A, B$. Order channels retained only by $A$, then channels shared by $A$ and $B$, then channels retained only by $B$. For example, say $A$ retains 1, 3 and $B$ retains 2, 3. Order $A$-only channels (1), then shared channels (3), then $B$-only channels (2) to produce a final ordering: 1, 3, 2.

More generally, we denote the ordered nodes in the mrap path as $p = [n_1,n_2,n_3,\ldots]$. We first take all the channels that are retained only by the first node, which we denote $\tilde{n_1}$. The first node should only share channels with the second node, so this is equivalent to computing the set subtraction: 
\begin{equation}
    \tilde{n_1} = n_1 - n_2
\end{equation}

More generally, we denote channels ``unique'' to node $i$ as $\tilde{n_i}$. The ith node should only share channels with the previous $n_{i-1}$ and next $n_{i+1}$ nodes, so take the difference to find all channels ``unique'' to node $n_i$.
\begin{equation}
\label{eqn:channels-unique-to-node}
\tilde{n_i} = n_i - n_{i-1} - n_{i+1}
\end{equation}

Then, take all channels shared by the first and second nodes, $n_1 \cap n_2$. Then, take channels unique to the second node $\tilde{n_2}$. Then, take channels shared by the second and third nodes, $n_2 \cap n_3$. Continue this for all nodes in the path. The total ordering looks like the following

\begin{equation}
\label{eqn:channel-ordering}
\tilde{n_1} \cup (n_1 \cap n_2) \cup \tilde{n_2} \cup (n_2 \cap n_3) \cdots
\end{equation}

Once the path is exhausted, continue the ordering with nodes in the following path. Once all paths are exhausted, the ordering is complete. This channel ordering is summarized as a permutation matrix $\Pi$ in Algorithm \ref{alg:upscale}. See Figure \ref{fig:unified} (Step 3) or Figure A.\ref{fig:ordering} for examples.

\subsection{Step 4 - Reorder Weights}
\label{sec:reordering-weights}

The channel ordering from the previous step is then used to reorder channels in producers and consumers. First, the channel order directly determines output channel order for all producers in the segment. More formally, for a list of model weights $W$ and producer index $p_i$, we denote the channel reordering as $W[p_i] \Pi$. Second, for each consumer, reorder the consumer's input channels accordingly (\ie, $W[c_i] \Pi$). See the full algorithm in Algorithm \ref{alg:upscale}, description in Appendix \ref{sec:reordering-pipeline-example} or example in Figure \ref{fig:unified} (Step 4).

\subsection{Generalized Method}
\label{sec:edge-cases}

\textbf{Some nodes are subsets of other nodes}. For example, consumer $A$ retains 1, 2, 3; consumer $B$ retains 1, 2; and consumer $C$ retains 2, 3. $A$ is the \textit{parent} node and $B, C$ are the \textit{children} nodes. Even though $A, B, C$ form a cycle in our graph, we can achieve a zero memory copy solution that accommodates all three consumers. Simply order channels as 1, 2, 3. This contradicts our acyclic path requirement. To handle this, we introduce two modifications: (1) Parent-child edges are ignored when rejecting cyclic paths, and (2) If child nodes in the path include all parent node channels, then the parent node's reward is added to the path's reward. For an extended description of this subset case and its resolution, see Appendix \ref{sec:node-subsets}.

\textbf{Multiple producers}. Our algorithm above assumes there is only one producer. To handle multiple producers, we find which channels correspond to one another, across producers. Say producer $A$ and $B$'s outputs are simply summed. In this case, channel 1 from $A$ is equivalent to channel 1 from $B$. Channel 2 from $A$ is equivalent to channel 2 from $B$, and so on and so forth. Knowing this, we can reduce to the single producer case, as an ordering for $A$'s filters automatically provides the ordering for $B$'s filters.

We can then run our method assuming convolution $A$ is the only producer. Afterwards, when reordering the producer weights (\ie, step 4), we map the original permutation $\Pi$ to each producer via the aforementioned equivalent channel map $\Pi_i$. Then order the weights by $W[p_i] \Pi_i \Pi$.
For a more detailed description and examples, see Appendix \ref{sec:reduction-to-single-producer}.

\section{Experiments}
\label{sec:experiments}

We present extensive experimentation to show that unconstrained pruning can attain significantly higher accuracy than constrained pruning, especially for modern, larger models and for those with complex topologies. For these unconstrained pruned models, we then show that \Ours~outperforms a baseline export in inference-time latency -- in fact, without \Ours, latency of the pruned models actually \textit{increases}, making \Ours~necessary for export. All prior methods utilize constrained pruning, so for our experiments, we remove those constraints and instead use \Ours~to export.

\begin{table*}[!ht]
    \centering
    \scriptsize
    \begin{tabular}{l|l|cccccccccc}
    \toprule
    Model & Heuristic & Statistic & 1\% & 5\% & 10\% & 15\% & 20\% & 25\% & 30\% \\
    \midrule
Densenet121 & HRank & Acc (Ours) & 72.04\% & 63.59\% & 40.68\% & 17.95\% & 4.87\% & 1.59\% & 0.62\% \\
& & Acc (Cons) & 72.36\% & 54.10\% & 17.03\% & 11.95\% & 1.46\% & 0.47\% & 0.11\% \\
& & Lat (Ours) & 5.44 ± 0.018 & 5.67 ± 0.016 & 5.69 ± 0.010 & 5.49 ± 0.005 & 5.41 ± 0.008 & 5.29 ± 0.005 & 5.63 ± 0.0085 \\
& & Lat (Base) & 5.93 ± 0.007 & 6.13 ± 0.004 & 6.15 ± 0.017 & 5.94 ± 0.011 & 5.89 ± 0.013 & 5.76 ± 0.008 & 6.11 ± 0.008 \\
& & Lat (Zero) & 5.29 ± 0.012 & 5.25 ± 0.003 & 5.21 ± 0.008 & 4.98 ± 0.004 & 4.93 ± 0.005 & 4.79 ± 0.003 & 5.59 ± 0.004  \\
    \midrule
    Squeezenet1\_1 & L2 & Acc (Ours) & 57.71\% & 39.84\% & 9.44\% & 0.88\% & 0.35\% & 0.19\% & 0.13\% \\
& & Acc (Cons) & 57.71\% & 29.70\% & 5.80\% & 1.41\% & 0.32\% & 0.12\% & 0.14\% \\
& & Lat (Ours) & 0.67 ± 0.003 & 0.67 ± 0.001 & 0.67 ± 0.002 & 0.67 ± 0.002 & 0.66 ± 0.002 & 0.65 ± 0.003 & 0.65 ± 0.002 \\
& & Lat (Base) & 0.73 ± 0.003 & 0.81 ± 0.001 & 0.84 ± 0.002 & 0.85 ± 0.002 & 0.82 ± 0.001 & 0.82 ± 0.003 & 0.81 ± 0.003 \\
& & Lat (Zero) & 0.67 ± 0.002 & 0.68 ± 0.001 & 0.67 ± 0.003 & 0.67 ± 0.001 & 0.67 ± 0.004 & 0.66 ± 0.001 & 0.65 ± 0.001 \\
\midrule
Mobilenet\_v3\_large & L1 & Acc (Ours) & 73.72\% & 64.88\% & 39.87\% & 2.23\% & 0.49\% & 0.16\% & 0.21\% \\
& & Acc (Cons) & 73.69\% & 52.93\% & 22.81\% & 1.06\% & 0.14\% & 0.18\% & 0.15\% \\
& & Lat (Ours) & 1.80 ± 0.005 & 1.75 ± 0.004 & 1.78 ± 0.005 & 1.75 ± 0.001 & 1.71 ± 0.004 & 1.69 ± 0.001 & 1.58 ± 0.003 \\
& & Lat (Base) & 1.77 ± 0.004 & 1.88 ± 0.008 & 1.89 ± 0.005 & 1.86 ± 0.003 & 1.82 ± 0.004 & 1.84 ± 0.003 & 1.84 ± 0.003 \\
& & Lat (Zero) & 1.56 ± 0.001 & 1.55 ± 0.002 & 1.49 ± 0.006 & 1.47 ± 0.002 & 1.45 ± 0.002 & 1.48 ± 0.003 & 1.43 ± 0.003 \\
    \bottomrule
    \end{tabular}
    \caption{ \textbf{Unconstrained vs. Constrained Accuracy and Latency} after applying post-training pruning -- experiments on various heuristics and architectures, across sparsity levels. Unconstrained accuracy matches or outperforms the baseline constrained accuracy. See full results across architectures, heuristics, and sparsity levels in Tables A.\ref{tab:upscale_vs_naive_latency_0}, A.\ref{tab:upscale_vs_naive_latency_1}, A.\ref{tab:upscale_vs_naive_latency_2}, A.\ref{tab:upscale_vs_naive_latency_3}.}
    \label{tab:efficientnet_accuracy_comparison} \color{white}{Easter Egg \#2!}
\end{table*}

\begin{table}[!ht]
    \centering
    \scriptsize
    \begin{tabular}{l|l|cccccccccc}
    \toprule
    Model & Type & 1\% & 10\% & 20\% & 30\% & 40\% \\
    \midrule
Efficientnet\_b1 
& Ours & 76.2\% & 74.4\% & 66.4\% & 46.0\% & 7.1\% \\
& Cons & 75.9\% & 72.0\% & 58.8\% & 24.4\% & 0.7\% \\
\midrule
Efficientnet\_b3 
& Ours & 78.0\% & 76.8\% & 69.8\% & 53.0\% & 15.2\% \\
& Cons & 77.7\% & 73.3\% & 60.1\% & 20.0\% & 0.4\% \\
\midrule
Efficientnet\_v2\_s 
& Ours & 80.3\% & 78.7\% & 65.6\% & 3.8\% & 0.3\% \\
& Cons & 80.0\% & 11.4\% & 0.1\% & 0.1\% & 0.1\% \\
    \bottomrule
    \end{tabular}
    \caption{ \textbf{Unconstrained vs. Constrained ImageNet Accuracy} after applying post-training pruning, across sparsity levels. Experiments above all use the LAMP heuristic. Unconstrained (\textit{Ours}) accuracy outperforms baseline constrained (\textit{Cons}) accuracy.}
\end{table}

\textbf{For the post-training setting, unconstrained pruning improves ImageNet top-1 accuracy by up to 76.7 points} over constrained pruning. Our goal is to assess the accuracy impact of switching from constrained to unconstrained pruning. For simplicity, we naively adapt pruning algorithms previously used for constrained pruning, to the unconstrained setting, by removing constraints on the pruning zero-one masks. To evaluate the effect of constrained or unconstrained pruning, independent of training recipes, we conduct experiments without fine-tuning, dubbed post-training pruning: (1) Take models pretrained on ImageNet, (2) apply various pruning heuristics at different sparsity levels, and (3) measure ImageNet top-1 validation accuracy. We sparsify parameters at intervals of 2.5\% from 0\% to 100\% and test 5 pruning strategies across 15+ architectures.

Although these pruning strategies were designed for constrained pruning, we find that unconstrained pruning achieves comparable or better accuracy than constrained pruning, for an average 2.1 point win averaged across all settings. For several cases, especially for complex topologies and larger models (Figure \ref{fig:unconstrained-larger-models}), unconstrained pruning yields significant accuracy benefits, up to a 21.7-point (DenseNet121, L1) increase in ImageNet accuracy, averaged across all sparsity levels -- or, up to a 76.7-point increase at a specific sparsity level (EfficientNetV2-Small, LAMP, 12.5\%). This demonstrates that unconstrained pruning can provide outsized benefits in the appropriate settings. We summarize results in Figure \ref{fig:unconstrained-vs-constrained-accuracy-main} and report full results in Figures A.\ref{fig:unconstrained_vs_constrained_accuracy_1}, A.\ref{fig:unconstrained_vs_constrained_accuracy_2}. Preliminary results for fine-tuning also show a sizable (5-point) accuracy gap per Appendix \ref{sec:fine-tuning}, and more thorough investigation is left to future work.

\textbf{Latency improves by up to 52.8\% when exporting pruned models using \Ours}, when compared with a baseline export for unconstrained pruning patterns -- this baseline is described in Section \ref{sec:why-memory-copies-occur}. To evaluate the effectiveness of our reordering algorithm independently, we (1) apply post-training pruning to models pretrained on ImageNet; (2) export unconstrained-pruned models with and without \Ours; and (3) benchmark the exported model's latency. \Ours~reduces latency of the exported model by 8.6\% on average across all settings and yields significant latency benefits, by up to 24.9\% (SqueezeNet1-1, L1), averaged across all sparsity levels -- or, up to a 52.8\% latency reduction at a specific sparsity level (ResNet18, FPGM). Critically, exporting unconstrained-pruned models without \Ours~actually \textit{increases} latency relative to the original, unpruned model; in the same setting, \Ours~is able to realize more appropriate latency reductions.

We summarize results in Figures \ref{fig:reordering_latency} and \ref{fig:unconstrained-larger-models}, reporting full results in Figure A.\ref{fig:unconstrained_vs_constrained_latency}. Note that there are no hyperparameters in our algorithm that can control performance, so we run our algorithm homogenously on all models to obtain said performance. We additionally plot latency for a theoretical zero-memory-copy solution, illustrating the maximum latency reduction attainable by any unconstrained pruning export; we observe that \Ours~often performs near-optimally, with latency nearly matching zero-memory-copy latency.

\textbf{Setup}. We use a single V100 GPU with 32 GB RAM. To export models for timing, we run an existing pruning strategy on the provided model, export using \Ours, then use PyTorch's jit trace to produce a Python-less executable. This traced model is then benchmarked using PyTorch's built-in profiling utility, including CUDA ``activities'' and tracking tensor memory allocation. Note this utility handles warmup automatically e.g., running several forward passes before initiating timed runs. All our latency measurements are the aggregate of 100 runs, with both mean and standard deviations reported. All accuracies are reported on the ImageNet ILSVRC 2015 \cite{russakovsky2015imagenet} validation dataset.

\section{Conclusion}

We introduce \OursFull~(\Ours)~to support pruning export more generically; out of the box, \Ours~handles a predominant challenge for pruning modern neural networks: namely, memory and thus latency inefficiency. Furthermore, we introduce both a framework and an approximate solution to mitigating inefficiencies, by reducing memory copies at inference time. The end result -- a generic pruning export library -- expands the total surface area that existing and new pruning algorithms can operate on, by allowing any pruning pattern to be exported and making unconstrained pruning a competitive alternative to conventional, constrained pruning.

\bibliography{upscale}
\bibliographystyle{icml2023}

\newpage
\appendix
\onecolumn

\begin{figure}
\includegraphics[width=\textwidth]{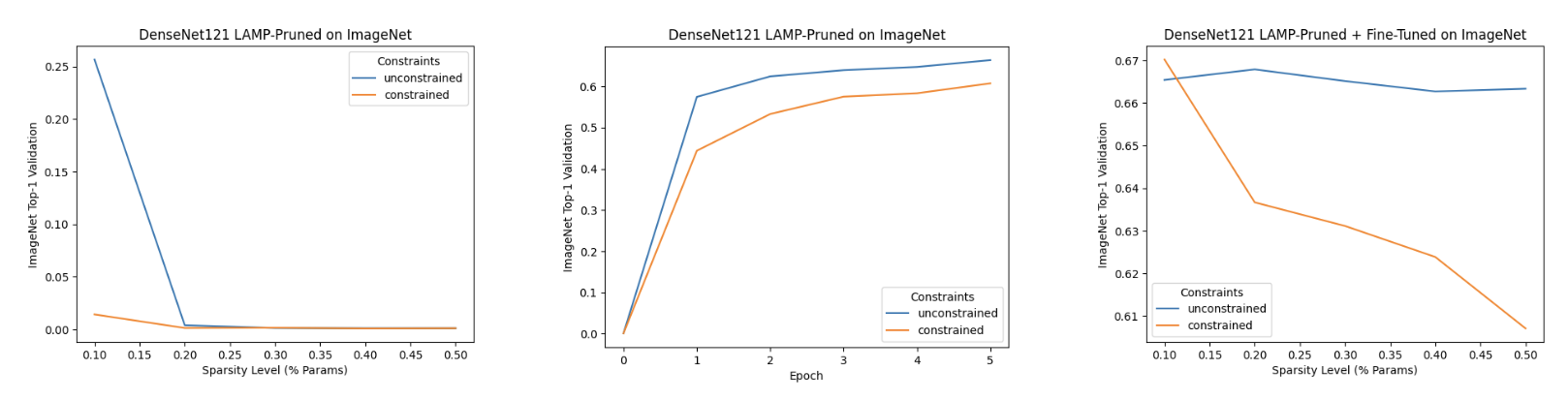}
    \caption{On the left, we show post-training pruned accuracy, with no fine-tuning. In the middle, we show only sparsity level 50\% but for fine-tuned accuracies over each epoch. This shows that the fine-tuned models have mostly converged. On the right, we show accuracies across sparsity levels, after fine-tuning for 5 epochs. In short, unconstrained still outperforms constrained after fine-tuning.}
    \label{fig:densenet121_ft}
\end{figure}

\section{How Fine-Tuning Affects Constraint Impact on Accuracy}
\label{sec:fine-tuning}

We note that the gap between unconstrained and constrained pruning remains significant ($>5$ points on ImageNet) even after fine-tuning.

We channel prune DenseNet121 at 10\%, 20\%, 30\%, 40\%, 50\% parameter sparsity using the LAMP heuristic — using both the constrained variant and an unconstrained variant. We then fine-tune all 10 models for 5 epochs each. These experiments show that unconstrained accuracy still significantly outperforms constrained accuracy on ImageNet. We show results in Figure \ref{fig:densenet121_ft}, from which we can make the following conclusions:

\begin{enumerate}
    \item After fine-tuning, unconstrained outperforms constrained by up to 6 ImageNet validation points.
    \item Unconstrained accuracy is almost independent of the pruning level, suggesting that constraints may hinder fine-tuned accuracy.
    \item We observe throughout training that unconstrained accuracy converges much quicker than constrained accuracy, so removing constraints can lower fine-tuning cost.
\end{enumerate}

Note that, the accuracy of the pruned model depends on many factors, including the dataset/task, pruning strategy, training strategy, model structure, etc. Therefore, it is hard to justify if unconstrained pruning always outperforms constrained pruning in terms of accuracy. Our major contribution was to provide a method/tool (\Ours) that is able to speed up any pruning methods (i.e., to bridge the latency gap between constrained pruning and unconstrained pruning). With the proposed \Ours, when designing a pruning method, we will only need to focus on the accuracy without worrying the latency regression brought by memory copies.

\section{How Reordering Reduces Memory Copies}
\label{sec:reordering-examples}

\begin{figure}
\centering
    \begin{minipage}{.48\textwidth}
    \centering
    \includegraphics[width=\textwidth]{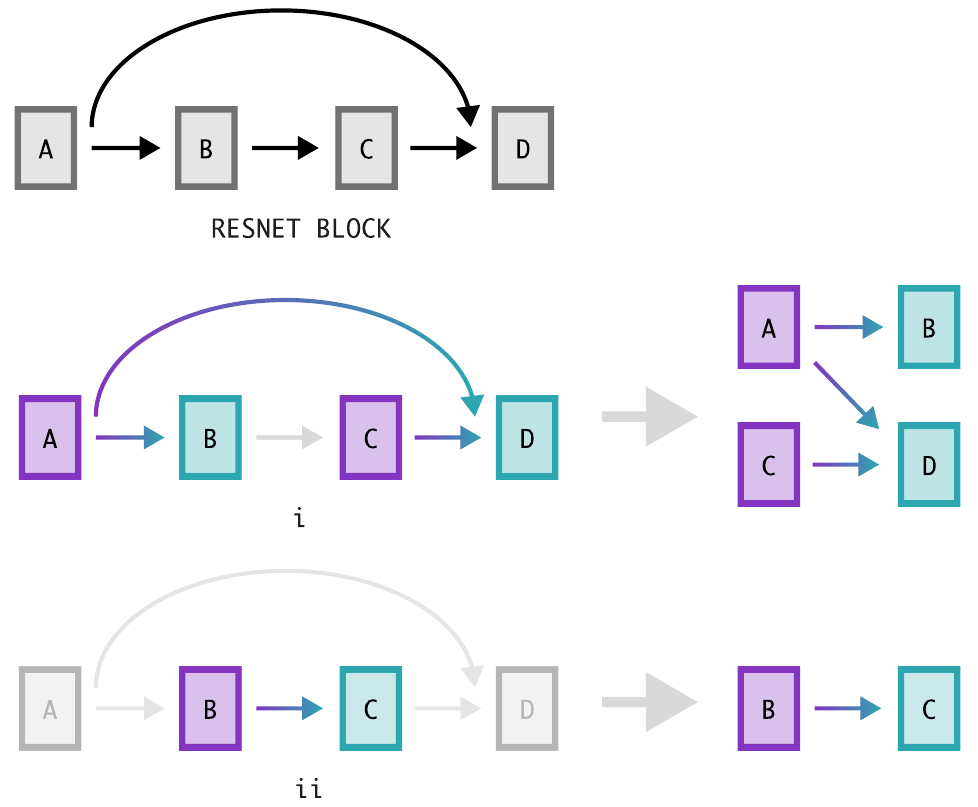}
    \caption{\textbf{Step 1 - Divide architecture into segments that can be independently pruned.} At the top, we illustrate a ResNet building block, with convolutions $A, B, C, D$. This block contains two segments, illustrated as i. in the middle row and ii. in the bottom row. Within each segment, a convolution is either a producer that produces output (purple) or a consumer that consumes input (blue). This is better illustrated in the ``canonical'' figures on the right, which places all producers on the left and consumers on the right.}
    \label{fig:segments}
\end{minipage}\hfill
    \begin{minipage}{.48\textwidth}
    \centering
    \includegraphics[width=\textwidth]{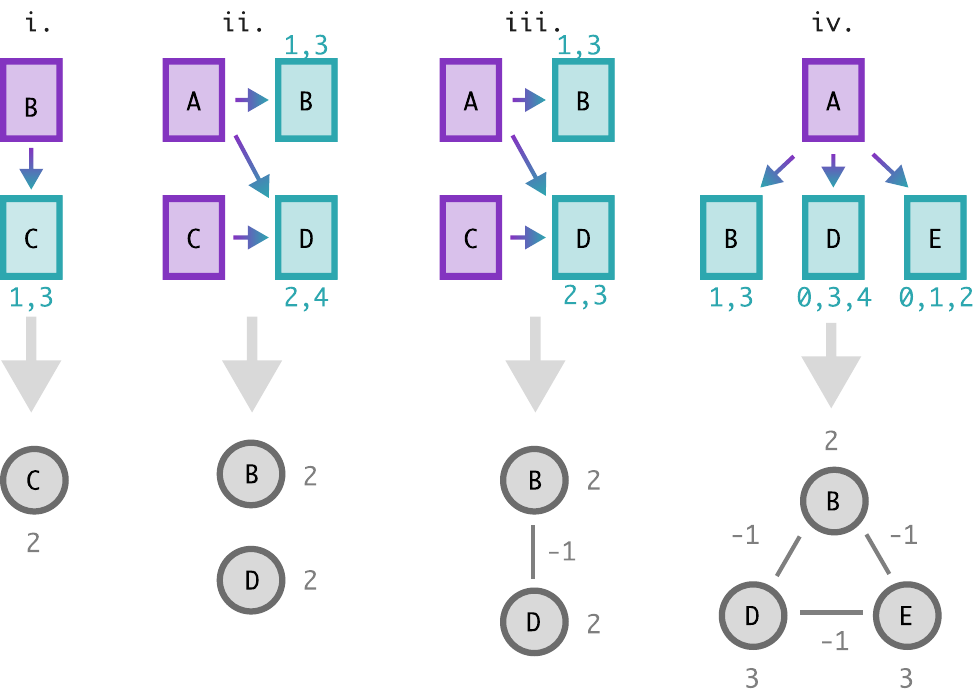}
    \caption{\textbf{Step 2 - Translate each segment into a graph}. Only consumers (blue box) are translated into nodes (gray circles). The blue numbers (ex. 1,3 in i.) are the channels each consumer retains. The gray numbers (e.g., 2 in i.) are the node and edge reward. Examples i, ii, iii are ResNet segments from Figure \ref{fig:segments}. (i) Since C retains two channels, its node has a reward of 2. (ii) B retains 1,3 and D retains 2,4; since both nodes retain two channels, their reward are each 2. They do not share channels so do not share an edge. (iii) B retains 1,3 and D retains 2,3; since both nodes retain channel 2, they share an edge with reward -1. (iv.) All nodes are densely connected, because every convolution shares channels with all other nodes.}
    \label{fig:graphs}
    \end{minipage}
\end{figure}

There are several cases, with various degrees of complexity, for consumer interactions. In each scenario, we can use reordering to minimize the number of memory copies. See Figure A.\ref{fig:segments} for the architecture.

\begin{enumerate}
\item \textbf{Case 1: Disjoint sets}. Say convolution $B$ retains channels 1 and 3. Convolution $C$ retains channels 2 and 4. To minimize memory copies, ensure each convolution's retained channels are side-by-side in the output tensor $X$. To do so, we rearrange the filters in convolution $A$ to be 1, 3, 2, 4. This means the channels in $X$ are also rearranged to be 1, 3, 2, 4. Notice the channels $B$ retains -- 1, 3 -- are now contiguous in memory in the first two channels of $X$. Channels $C$ retains -- 2, 4 -- are contiguous in the last two channels of $X$. Now, no memory copies are needed at inference time, as convolution $B$ takes the first two channels $X[:2]$ and $C$ takes the last two $X[2:]$. (Figure \ref{fig:unified}, Case 1)
\item \textbf{Case 2: Overlapping sets}. Say convolution $B$ retains channels 1 and 3. Convolution $C$ retains channels 2 and 3. Again, ensure each convolution's retained channels are side-by-side in $X$. To do so, rearrange the filters in convolution $A$ to be 1, 3, 2. More generally, include channels unique to $B$, then channels shared by $B$ and $C$, then channels unique to $C$. Now, the channels $B$ retains -- 1, 3 -- are contiguous and channels $C$ retains -- 2, 3 -- are contiguous. However, for convolution $C$, notice $X$ contains channels in the reverse order: 3, 2 in stead of 2, 3. As a result, we \textit{also} need to reorder input channels in convolution $C$ to be 3, 2. Now, no memory copies are needed at inference time, as $B$ uses $X[:2]$ as input and $C$ uses $X[2:]$. (Figure \ref{fig:unified}, Case 2)
\item \textbf{Case 3: Unsolvable sets}. Say convolution $B$ retains channels 1, 3, 4. Convolution $C$ retains channels 2, 3, 4. We introduce another convolution $D$, which retains channels 1, 2. There is no ordering of channels that would completely eliminate memory copies, as satisfying any pair of convolutions would necessarily exclude the remaining convolution. We will formalize the definition of this unsolvable case later. For now, our intuition is to instead maximize the number of channels placed in the correct order. This means satisfying convolutions $B$ and $C$ and ignoring convolution $D$, since convolutions $B$ and $C$ each retain 3 channels and convolution $D$ retains only 2 channels.
\end{enumerate}

\begin{figure}[!ht]
\centering
    \begin{minipage}{.48\textwidth}
        \centering
        \includegraphics[width=\textwidth]{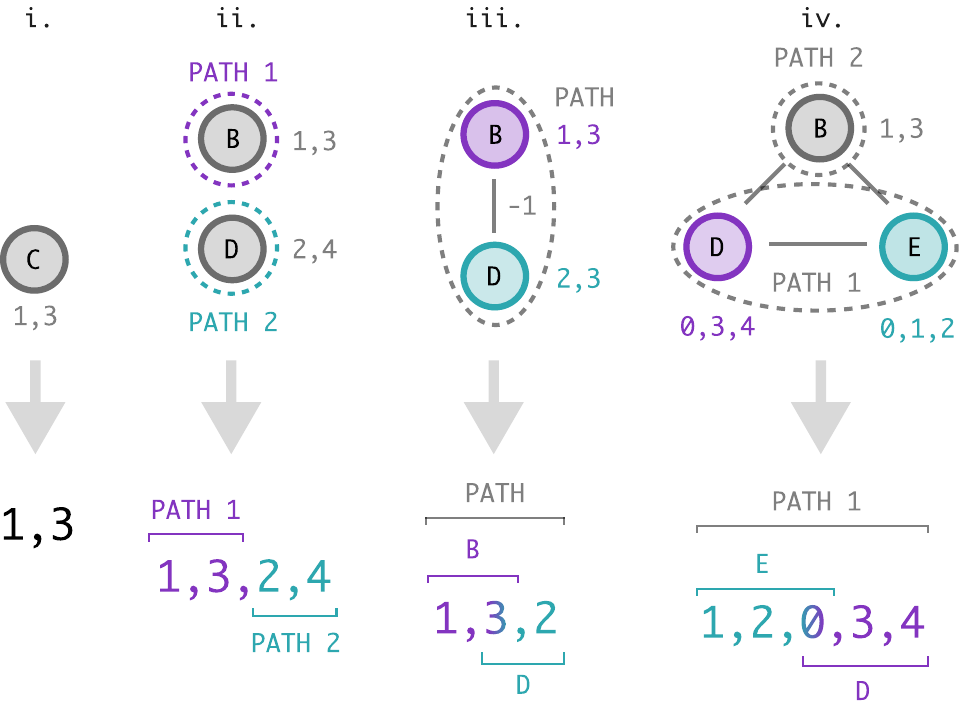}
        \caption{\textbf{Step 3 - Translate graph into channel ordering}. Find maximum reward acyclic paths. Then, translate paths into a channel ordering. Examples i, ii, iii are ResNet segments from Figure \ref{fig:segments}. (i) There is only one node, so order its channels arbitrarily. (ii) Since there are no edges, there are two single-node paths. Order path 1 node's channels (purple), then path 2 node's channels (blue). (iii) There is one path with both nodes. Order B-only channels (1, purple), B-and-D shared channels (3, purple-blue), then D-only channels (2, blue). (iv) Since all three nodes are densely connected, the first path can only contain two of three nodes. The two nodes with the highest reward are D and E, each with reward three, so they comprise path 1. B comprises path 2. Per path 1, take E-only channels (1, 2, blue), then E-D shared channels (0, blue-purple), then D-only channels (3, 4, purple). All channels in the second path are already covered by the ordering, so the ordering is complete.}
        \label{fig:ordering}
    \end{minipage}\hfill
    \begin{minipage}{.48\textwidth}
        \centering
        \includegraphics[width=\textwidth]{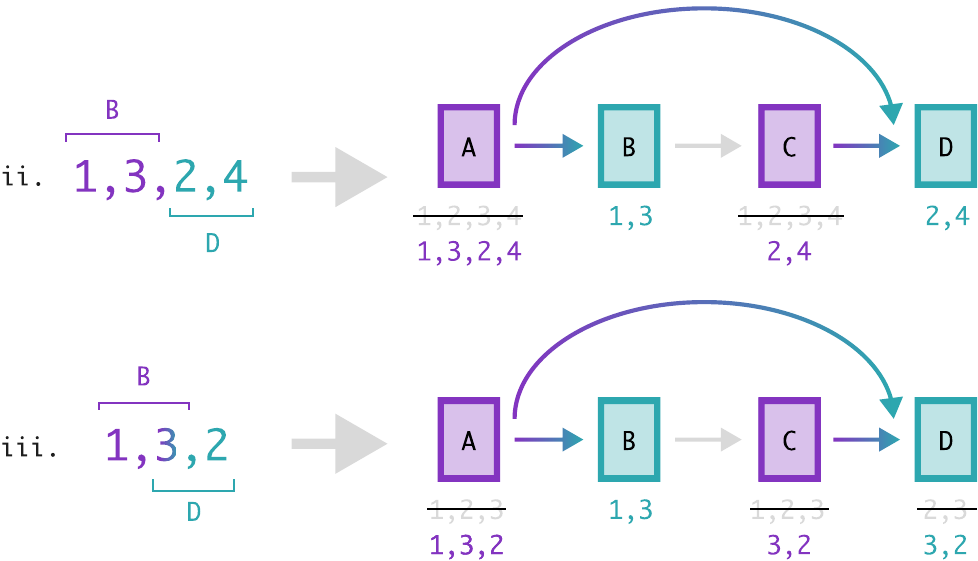}
        \caption{\textbf{Step 4 - Reorder weights.} ii, iii correspond to ii, iii in previous figures. (ii) The final ordering dictates the ordering for producer $A$ and $C$. However, producer $C$ omits 1,3 as none of its consumers ($D$) use those channels. Consumers $B$ and $D$ do not need reordering, as their retained channels 1,3 and 2,4 were not reordered. (iii) Consumer $B$ does not need reordering, as the final ordering lists its channels 1,3 in order. However, consumer $D$ needs to reorder its channels from 2,3 to 3,2.}
        \label{fig:step4}
    \end{minipage}
\end{figure}

\section{How Graph Algorithm Solves Reordering}
\label{sec:graph-algorithm-examples}

For the three cases in the previous section, Appendix \ref{sec:reordering-examples}, here is how our graph would be constructed and how the solution path minimizes memory copies (Figures \ref{fig:segments},\ref{fig:graphs},\ref{fig:ordering},\ref{fig:step4}).

\begin{enumerate}
\item \textbf{Case 1: Disjoint Sets}. In the simplest case, no convolutions share retained channels, so when rearranging channels, place all channels retained by one convolution together, then all channels retained by a second convolution etc. Any ordering of convolutions works. Translated into our graph: There are no edges, and we can pick any ordering of nodes. The ordering of nodes then directly corresponds to an ordering of channels.
\item \textbf{Case 2: Overlapping Sets}. Say we have a linear ordering of convolutions, where each convolution shares retained channels only with the preceding and succeeding convolutions. In this case, there always exists a zero-copy solution: start with channels unique to the first convolution, channels shared between the first and second, channels unique to the second convolution, channels shared between the second and third etc. Translated into our graph: Find a set of nodes where each node has at most two neighbors -- in other words, an acyclic path.
\item \textbf{Case 3: Unsolvable Sets}. Say one or more convolutions share retained channels with all other convolutions. This means we have a cycle: Attempting to order the nodes linearly will result in one node belonging at both the end and the start at the same time. In other words, we cannot achieve a perfect zero-memory-copy solution. In this case, our goal is to maximize the number of channels that don't need to be copied. Translated into our graph: Assign a reward to every node and edge. Every node's reward is the number of channels the corresponding convolutions retains. Every edge's reward is \textit{negative} of the number of shared channels between the two node's convolutions. This way, a path's reward corresponds to the number of channels included in that path. To maximize the number of copy-free channels, minimize the path reward. In short, find the maximum reward acyclic path.
\end{enumerate}

\begin{figure}[!ht]
\centering
    \begin{minipage}{.48\textwidth}
    \centering
    \includegraphics[width=\textwidth]{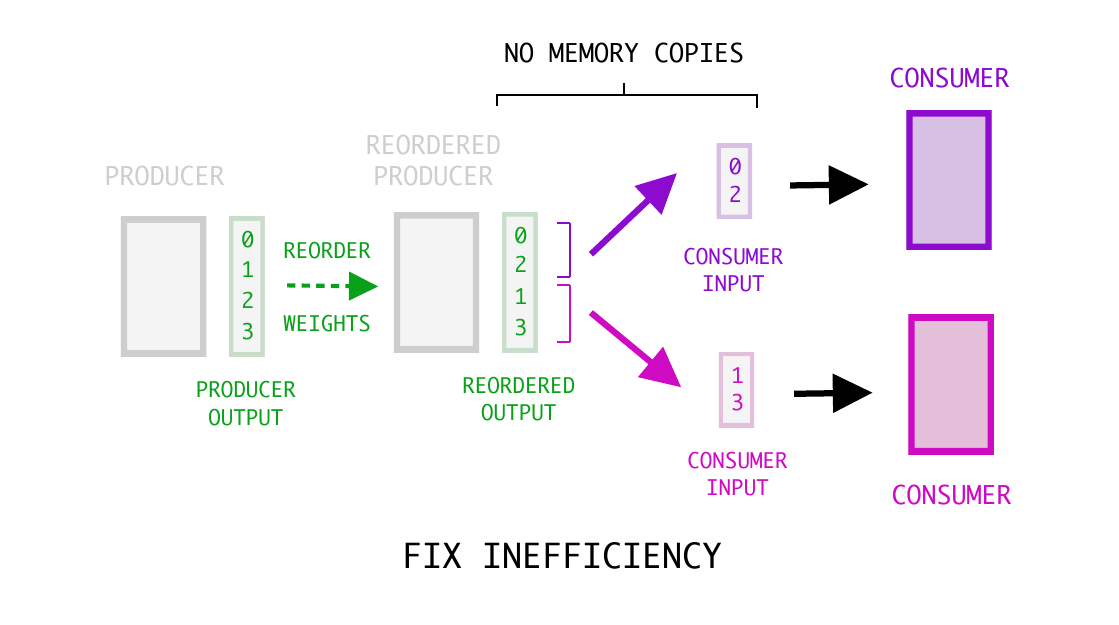}
    \caption{\textbf{How to fix inefficient memory copies for pruned multi-branch segments?} To avoid memory copies at inference time reorder producer weights as shown in green above. This ensures that the two consumers (purple and pink) take as input contiguous slices of the green output tensor. Since these slices are already contiguous in memory, no memory copies are needed to assemble the pair of consumer inputs.}
    \label{fig:fix-inefficiency}
   \end{minipage}\hfill
    \begin{minipage}{.48\textwidth}
    \centering
    \includegraphics[width=\textwidth]{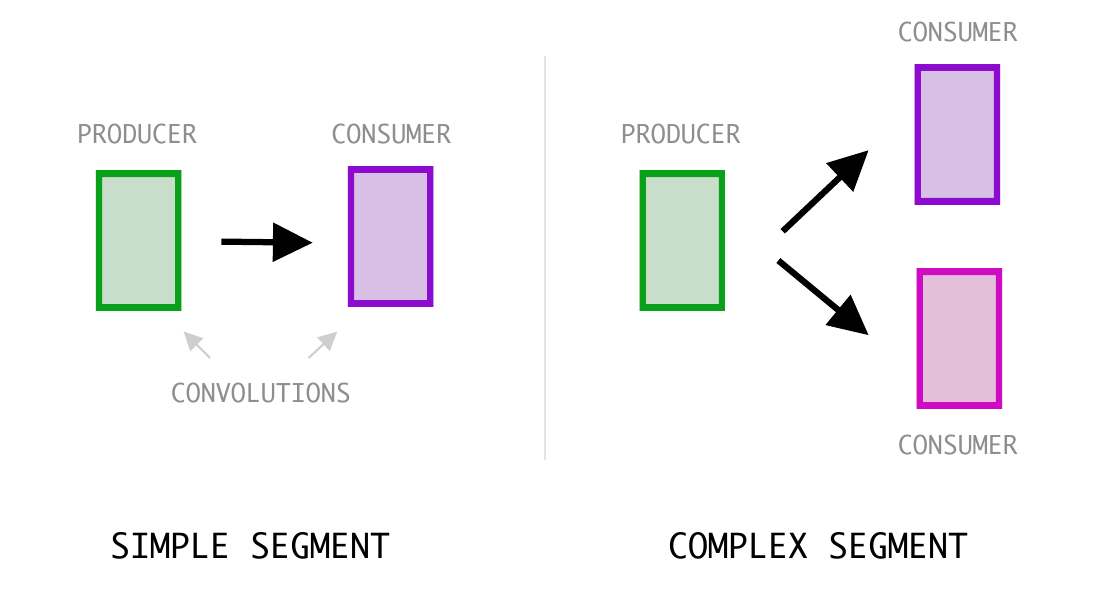}
    \caption{ A \textbf{segment} is made up of (a) convolutions that have their outputs combined (e.g., summed or concatenated), along with (b) all the convolutions that use those values. We call the convolutions producing output \textit{producers} and the convolutions consuming values \textit{consumers}. We define a single-branch segment (left) to be a part of the architecture that includes just one convolution producing output and one convolution consuming input. We define a multi-branch segment (right) to be a part of the architecture that contains \textit{multiple} convolutions producing outputs or \textit{multiple} convolutions consuming inputs.}
    \label{fig:simple-vs-complex-segments}
    \end{minipage}
\end{figure}

\section{Handling Node Subsets}
\label{sec:node-subsets}

Recall each node is equivalent to a convolution. In particular, each node represents the set of retained channels for a convolution. Some nodes may be strict sets of another node: For example, one node $A$ may correspond to retained channels 1, 2, 3. Call this the \textit{parent}. However, two other nodes may correspond to channels 1, 2 ($B$); and 2, 3 ($C$). Call these the \textit{child} nodes. Although there exists a cycle between $A, B, C$, there exists a valid channel ordering: 1, 2, 3.

As a result, nodes with parent-child relationships break the requirement for an acyclic path. One possibility is to simply drop parent or child nodes, to ensure retain the acyclic path requirement. However, parent node neighbors may influence within-parent channel ordering, which conflicts with the ordering imposed by child nodes. Here are two examples:
\begin{enumerate}
\item \textbf{Child nodes cannot be ignored}. One natural idea is to include only the parent node in the graph and drop all child nodes. Here is a counterexample: Say there are nodes $A$, $B$, $C$. $A$ retains 1, 2, 3. $B$ retains 2, 3, 4, 5. $C$ retains 4, 5, 6. The ordering is straightforward: $A \rightarrow B \rightarrow C$, inducing the order 1, 2, 3, 4, 5, 6. However, say there also exists child nodes $D$, which retains 3, 4, 5 and $E$, which retains 2, 3, 5. By ignoring $D, E$ and using $B$, we incur more memory copies. Instead, the algorithm should including the two child nodes, ordering $A \rightarrow D \rightarrow E \rightarrow C$, inducing the order 1, 2, 3, 5, 4, 6.
\item \textbf{Parent nodes cannot be ignored: }. Another natural idea is to include only the child nodes in the graph and drop all parent nodes. However, parent nodes are only automatically accounted for if enough child nodes are used to cover all channels retained by the parent node. In any other case, the algorithm must choose between child nodes \textit{or} the parent node. For example, say we have nodes $A, B, C$, which retain 1, 2, 3, 4; 2, 3, 4, 5; and 4, 5, 6, respectively. Then, child nodes $D, E$ retain 2, 4, 5; and 2, 3, 5. There are 3 options here: $A \rightarrow B \rightarrow C$ or $A \rightarrow D \rightarrow C$ or $A \rightarrow E \rightarrow C$. The maximum reward path is actually the one that includes the parent node $B$.
\end{enumerate}

Knowing this, our only option left is to include both child and parent nodes in the graph. Instead, we introduce two modifications:

\begin{enumerate}
\item Permit cycles between parent and child nodes.
\item Modify the reward function.
\end{enumerate}

We can modify the reward functions in the following way, by breaking down all possible ways to incorporate parent-child nodes in the solution path.

\begin{itemize}
    \item If select child nodes after selecting parent
    \begin{itemize}
        \item Parent contributes to reward regardless of how you leave the parent's group of child nodes. (no modification needed)
    \end{itemize}
    \item If select child nodes without selecting parent
    \begin{itemize}
        \item Parent contributes to reward if you leave by selecting the parent (no modification needed)
        \item Alternatively, tour enough children nodes to cover the parent node's channels, to contribute to the reward.
    \end{itemize}
\end{itemize}

As a result of this breakdown, we only need to modify the reward if enough child nodes are toured to cover the parent node's channels. In other words: \textit{If included child nodes include all channels for the parent node, then the parent node's reward is automatically added to the path's reward. If otherwise, the parent node contributes to the reward normally, depending on its participation in the maximum-reward path.}

\section{Reduction to Single Producer}
\label{sec:reduction-to-single-producer}

To reduce the multiple-producer case to the single-producer case, we make two critical assumptions, for operations \textit{other} than a multiply-reduce:

\begin{enumerate}
    \item \textbf{Producer channels are not ``mixed'' with other channels from the same producer}. Here, channels $i$ and $j$ are ``mixed'' if they both affect one output channel. For example $y = x[:, i] + x[:, j]$ would ``mix'' both channels $i$ and $j$. Using this definition of ``mix'': our assumption is that channel $i$ from any producer is never mixed with channel $j \neq i$ from that same producer. This does not preclude channel $i$ from mixing with itself. This is generally a true assumption, as channel mixing in common architectures is parameterized as a multiply-reduce. Note that neither channel shuffling nor transformers would violate this assumption. 
    \item \textbf{If producer $A$ mixes with producer $B$, exactly one channel from $A$ mixes with exactly one channel from $B$.} We assume that producer outputs are mixed relatively simply without channel micro-management. This is generally a true statement in common architectures as well, as the only channel-related operations like \texttt{flatten} only increase the number of feature map channels associated with a producer channel.
\end{enumerate}

Given the two above assumptions, we can then establish an equivalence class of channels across producers. In particular, we can define a one-to-one mapping from producer $A$'s $i$th channel to producer $B$'s $j$th channel and repeat this for all channels. With this mapping constructed, we can then (1) reduce the multi-producer case to the single-producer case by replacing $B.j$ with $A.i$, (2) run the single-producer, multi-consumer algorithm, then (3) use the order of producer $A$'s channels to determine the ordering of producer $B$'s channels.

If we have a concatenation operation, as opposed to an addition for example, then we simply treat every producer independently, as there are no cross-producer constraints.

\section{Unconstrained Output Pruning}
\label{sec:unconstrained-output-pruning}

\begin{figure}
    \centering
    \includegraphics[width=.95\textwidth]{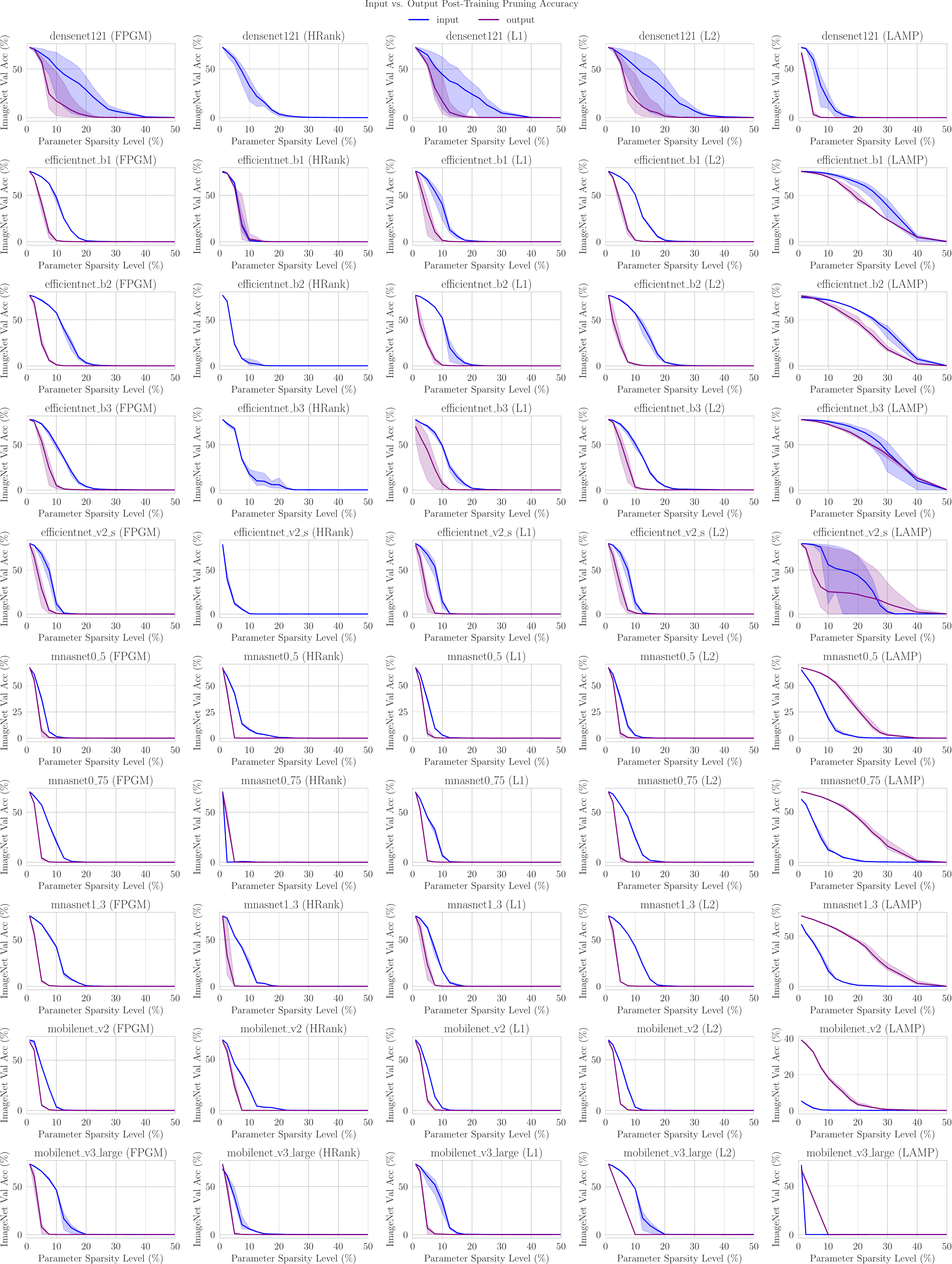}
    \caption{\textbf{Input vs. Output Pruning Accuracy} (Part 1) Input pruning (blue) outperforms output pruning (purple) in accuracy.}
    \label{fig:input_vs_output_accuracy_1}
\end{figure}

\begin{figure}
    \centering
    \includegraphics[width=.95\textwidth]{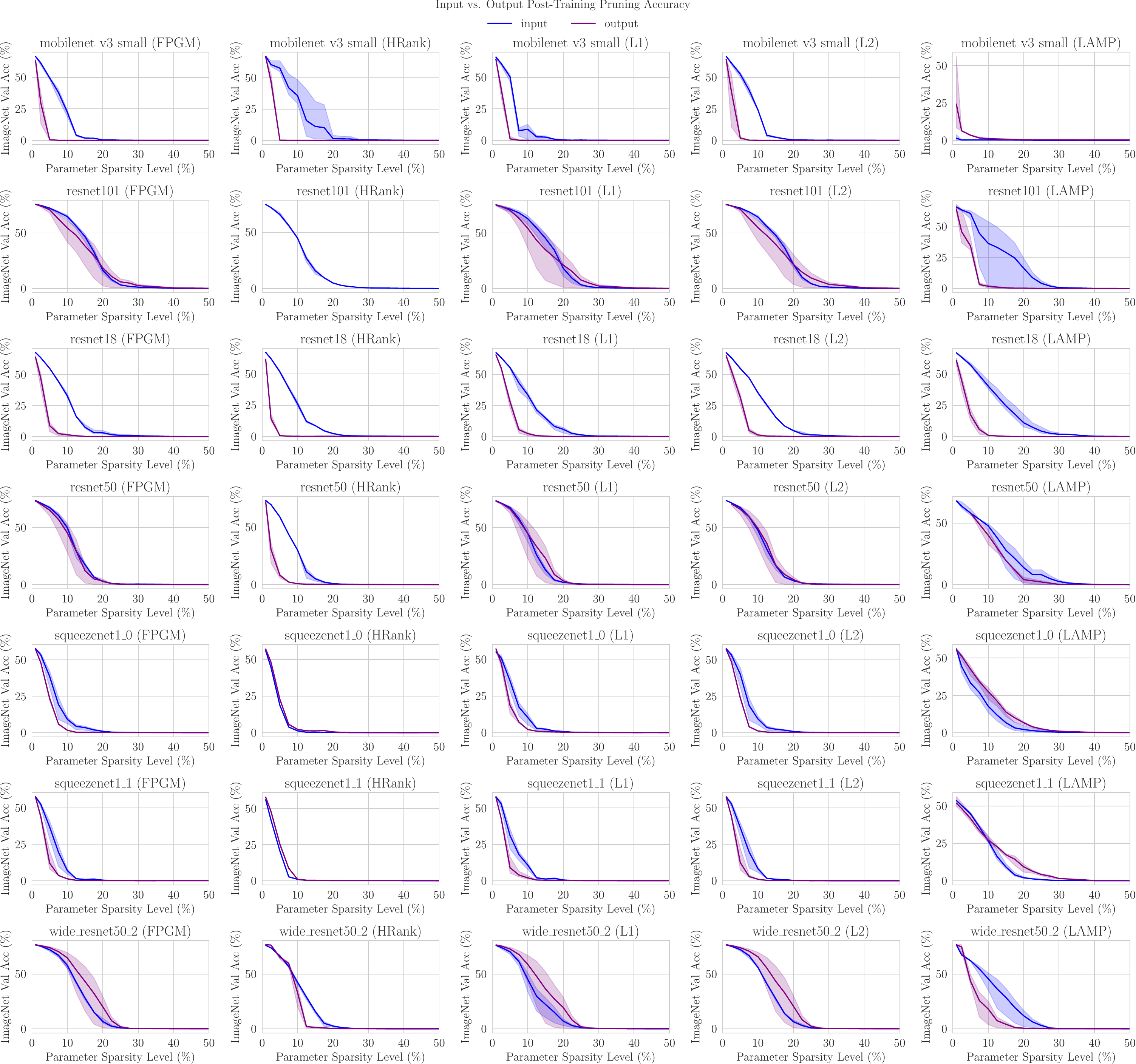}
    \caption{\textbf{Input vs. Output Pruning Accuracy} (Part 2) Input pruning (blue) outperforms output pruning (purple) in accuracy.}
    \label{fig:input_vs_output_accuracy_2}
\end{figure}

The main text describes unconstrained input pruning in detail but omits a description of the output pruning algorithm. We note empirically that input pruning outperforms output pruning (Figures \ref{fig:input_vs_output_accuracy_1}, \ref{fig:input_vs_output_accuracy_2}). There are several subtleties in exporting an unconstrained output-pruned model. Like before, memory copies are an issue, but rather than copy channels to produce input for the subsequent convolution, the memory copies now stem from assembling incompatible convolutional outputs.

Say we have two convolutions $A$ and $B$. Both outputs have 4 channels and are summed together. There are several possible pruning patterns and associated problems:

\begin{enumerate}
\item Convolution $A$ prunes channel 2. Now, the two outputs feature incompatible dimensions.
\item Convolution $A$ prunes channel 2, and convolution $B$ prunes channel 3. The two outputs feature compatible dimensions, but naively summing the two tensors is incorrect.
\end{enumerate}

To resolve this, the naive solution is to infill zeros, but this incurs a large number of memory copies. This infilling operation is the naive baseline for output pruning latency.

Instead, we can sum all channels that both convolutions $A$ and $B$ both retain. Then, we can concatenate with channels unique to $A$ and $B$ each. This is reminiscent of unconstrained input pruning: Reorder filters in $A$ and $B$ to place channels unique to $A$ first, then channels shared by $A$ and $B$, then channels unique to $B$. This is exactly how we order channels for unconstrained pruning.

To support unconstrained output pruning, we define each node in the graph to be a convolution \textit{producing} output -- as opposed to a convolution \textit{consuming} input, as in unconstrained input pruning. Then, add edges when those convolutions retain the same channels. The remainder of the algorithm remains the same.

\textit{Note that output pruning is sensitive to a correctness issue: ``pruned'' channels do not stay ``pruned'', as zero is not idempotent under operations such as a bias. This makes a general purpose export for output pruning slightly more tricky.}

\section{Reordering Pipeline Example}
\label{sec:reordering-pipeline-example}

Here is the full reordering pipeline. For a more formal description, see Algorithm \ref{alg:upscale} in the main text. Reuse the same example as above: There is one convolution $A$ producing 4 channels. Convolution $B$ retains channels 1, 3, 4; $C$ retains channels 2, 3, 4; and $D$ retains channels 1 and 4.
\begin{enumerate}
\item Build a graph: every node contains channels retained by the corresponding convolution. Create edges between nodes when their convolutions share retained channels. We have 3 nodes, labeled $B$, $C$, and $D$. All 3 nodes are densely connected.
\item Find the maximum reward acyclic path. Every node corresponds to a convolution, so the path admits a convolution ordering. In this case, our algorithm computes path $B \rightarrow C$, excluding $D$.
\item This convolution ordering admits a channel ordering: Start with channels unique to the first convolution, then channels shared between the first and second convolution, the channels unique to the second convolution etc. We then write the channel ordering: (1), (3, 4), (2). We use parentheses to denote channels unique to $B$, those shared by $B$ and $C$, then those unique to $C$.
\item Use this ordering to reorder filters in the convolution producing output. Output channels -- \ie, the filters -- in $A$ are rearranged as 1, 3, 4, 2.
\item Use this ordering to reorder filters in all convolutions consuming the reordered tensor. Input channels for $B$ are left in the same order, as 1, 3, 4. Input channels for $C$ are reordered to be 3, 4, 2.
\end{enumerate}

\begin{table*}
\centering
\footnotesize
\begin{tabular}{l|lllllllllll}
\toprule
Model &   Stat &    Segments &   10\% &   20\% &   30\% &   40\% &   50\% &   60\% &   70\% &   80\% &   90\% \\
\midrule
ResNet18             &  FLOPs &  All (ours) &  1.60 &  1.36 &  1.13 &  0.93 &  0.71 &  0.52 &  0.34 &  0.19 &  0.07 \\
ResNet18             &    Acc &  All (ours) & 54.88 & 51.94 & 51.00 & 49.75 & 45.24 & 39.94 & 31.07 & 19.34 & 10.27 \\
ResNet18             &  FLOPs &      Single-branch &  1.41 &  0.99 &  0.57 &  0.17 &  0.16 &  0.16 &  0.16 &  0.16 &  0.16 \\
ResNet18             &    Acc &      Single-branch & 54.50 & 49.87 & 41.61 &  8.77 &  8.28 &  7.49 &  8.38 &  8.64 &  9.23 \\
\midrule
MobileNetV3-Small    &  FLOPs &  All (ours) &  0.05 &  0.04 &  0.04 &  0.03 &  0.02 &  0.02 &  0.01 &  0.01 &  0.00 \\
MobileNetV3-Small    &    Acc &  All (ours) & 59.32 & 55.06 & 51.34 & 41.67 & 34.46 & 24.20 & 15.30 & 10.03 &  3.68 \\
MobileNetV3-Small    &  FLOPs &      Single-branch &  0.05 &  0.04 &  0.04 &  0.04 &  0.04 &  0.04 &  0.04 &  0.04 &  0.04 \\
MobileNetV3-Small    &    Acc &      Single-branch & 55.33 & 37.35 &  0.10 &  0.15 &  0.11 &  0.10 &  0.10 &  0.11 &  0.12 \\
\midrule
MobileNetV3-Large    &  FLOPs &  All (ours) &  0.19 &  0.16 &  0.13 &  0.11 &  0.08 &  0.06 &  0.04 &  0.02 &  0.01 \\
MobileNetV3-Large    &    Acc &  All (ours) & 64.23 & 62.82 & 59.81 & 56.55 & 49.79 & 41.03 & 27.30 & 18.75 &  8.14 \\
MobileNetV3-Large    &  FLOPs &      Single-branch &  0.18 &  0.15 &  0.14 &  0.14 &  0.14 &  0.14 &  0.14 &  0.14 &  0.14 \\
MobileNetV3-Large    &    Acc &      Single-branch & 63.84 & 55.08 &  0.13 &  0.14 &  0.16 &  0.16 &  0.14 &  0.16 &  0.16 \\
\midrule
EfficientNetV2-Small &  FLOPs &  All (ours) &  2.47 &  2.08 &  1.69 &  1.34 &  0.98 &  0.69 &  0.41 &  0.19 &  0.06 \\
EfficientNetV2-Small &    Acc &  All (ours) & 67.92 & 66.25 & 65.53 & 62.77 & 59.66 & 54.16 & 44.80 & 33.36 & 20.04 \\
EfficientNetV2-Small &  FLOPs &      Single-branch &  2.24 &  1.61 &  1.43 &  1.43 &  1.43 &  1.43 &  1.43 &  1.43 &  1.43 \\
EfficientNetV2-Small &    Acc &      Single-branch & 67.40 & 59.85 &  0.27 &  0.32 &  0.33 &  0.31 &  0.31 &  0.28 &  0.31 \\
\bottomrule
\end{tabular}
\caption{\textbf{Accuracy vs. FLOPs with Fine-Tuning}. ``Single-branch'' means we prune only single-branch segments in the network to attain the desired compression ratio. ``All'' means we pruned all segments -- single-branch and multi-branch -- to reach the desired compression ratio. Across all models, pruning all segments achieves a better accuracy-latency tradeoff curve.}
\label{tab:acc_vs_flops_ft}
\end{table*}

\begin{figure}
\centering
    \begin{minipage}{.48\textwidth}
        \centering
        \includegraphics[width=\textwidth]{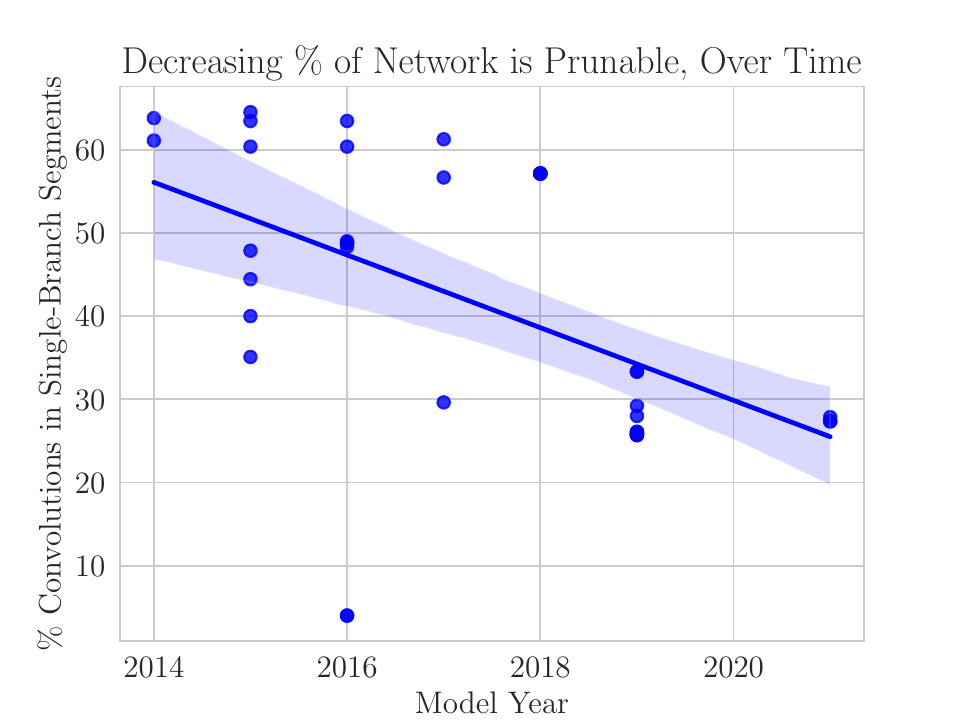}
        \caption{\textbf{Single-branch segments account for fewer and fewer of the convolutions over time}: Decreasing numbers of convolutions in neural network architectures belong to a single-branch segment, starting from 65\% with the advent of ResNets (2015) and decreasing to 26\% with models like EfficientNet (2019). Due to this, pruning export algorithms must accommodate multi-branch portions of the architecture.}
        \label{fig:prunable_over_time}
    \end{minipage}\hfill
    \begin{minipage}{.48\textwidth}
        \centering
        \includegraphics[width=\textwidth]{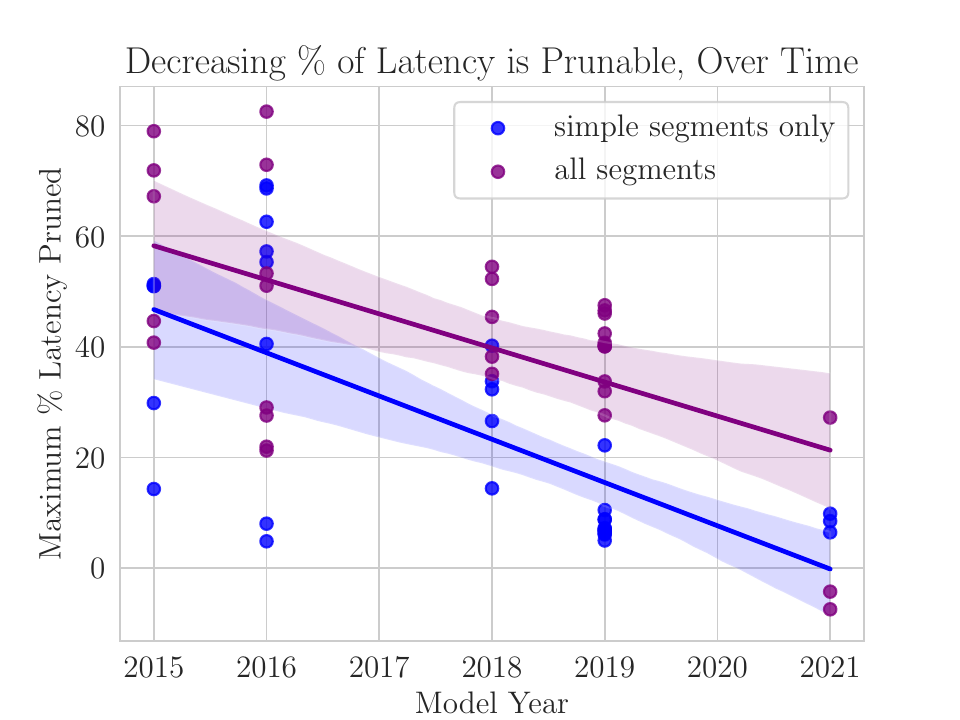}
        \caption{\textbf{Single-branch segments make up less and less of the total latency over time}: We prune the maximum number of channels in only single-branch segments, which prior methods can export; just as with FLOP count, single-branch methods can prune less and less of the total latency over time, only able to prune 4.9\% of the total latency for MobileNet-V3. By contrast, by pruning all the segments that \Ours~algorithm can export -- meaning all segments -- we are able to reduce a further 15\% of latency on average.}
        \label{fig:latency_reduction_over_time}
    \end{minipage}
\end{figure}

\begin{table}
    \centering
    \begin{minipage}[b]{.5\linewidth}
        \centering
        \small
        \begin{tabular}{l|lll}
            \toprule
            Model & Method & Latency & Accuracy \\
            \midrule
            ResNet18 & L2 & 3.74 & 54.5\% \\
            ResNet18 & \Ours & \textbf{3.45} & \textbf{54.9\%} \\
            \midrule
            ResNet18 & L2 & 3.60 & \textbf{49.9\%} \\
            ResNet18 & \Ours & \textbf{2.54} & 49.8\% \\
            \midrule
            MobileNetV3-Small & L2 & 1.84 & 37.3\% \\
            MobileNetV3-Small & \Ours & \textbf{1.79} & \textbf{51.4\%} \\
         \bottomrule
        \end{tabular}
        \caption{\textbf{Accuracy vs. Latency for Magnitude Pruning}: A summary of accuracy-latency tradeoffs between pruning simple segments only (L2) compared with pruning all segments (L2-\Ours). In the latter, since we additionally prune multi-branch segments, we use \Ours~to export the model for final latency measurements. Across various model sizes and sparsity levels, we find that models using \Ours~attain lower latency and comparable or higher accuracy.}
        \label{tab:acc_vs_latency_ft}
    \end{minipage}\hfill
    \begin{minipage}[b]{.47\linewidth}
        \includegraphics[width=\textwidth]{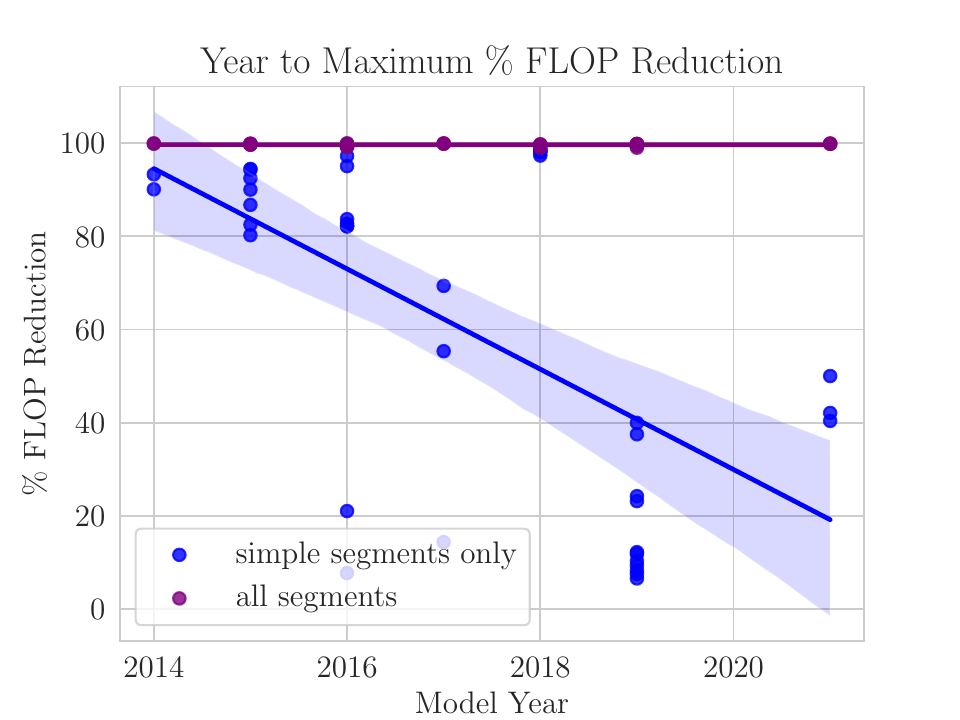}
        \caption{\textbf{Single-branch segments make up less and less of the FLOP count over time}: Across models over the past decade, single-branch segments make up precipitously less and less of the total FLOP count, only accounting for 6.5\% of the total computational cost for EfficientNet. By contrast, by pruning all the segments that our \Ours~algorithm can export -- meaning all segments -- we are able to prune 99.5\% of the computational cost, on average.} 
        \label{fig:flop_reduction_over_time}
    \end{minipage}
\end{table}

\section{Importance of Multi-Branch Segments}
\label{sec:multi-branch-segments}

One possible way to handle multi-branch segments is to completely ignore them. In this appendix section, we explore this possibility.

Unfortunately, this strategy is not viable: Over time, less and less computational complexity in a neural network is due to single-branch segments. More and more are involved in multi-branch segments of a neural network. This is for good reason, as this has significant impact on accuracy: Pruning 10\% of all weights is more difficult when you're restricted to just single-branch segments. We show this in Figures \ref{fig:prunable_over_time},\ref{fig:latency_reduction_over_time},\ref{fig:flop_reduction_over_time} and Tables \ref{tab:acc_vs_latency_ft},\ref{tab:acc_vs_flops_ft}.

\textbf{Single-branch segments account for as little as 4.9\% of latency, 6.5\% of FLOPs and 26\% of convolutions} for modern, efficient models. We could focus on train-time pruning performance for single-branch segments, but for compact, efficient models, only a small percentage of latency is due to single-branch segments. As a consequence, this approach is bottlenecked, being able to prune only a small percentage of latency in more modern neural networks--as little as 4.9\% of latency in MobileNetV3-Small (Figure \ref{fig:latency_reduction_over_time}), 6.5\% of FLOPs in EfficientNet-B7 (Figure \ref{fig:flop_reduction_over_time}), and 26\% of the convolutions in MobileNetV3-Large (Figure \ref{fig:prunable_over_time}).

\begin{figure}
    \centering
    \includegraphics[width=\textwidth]{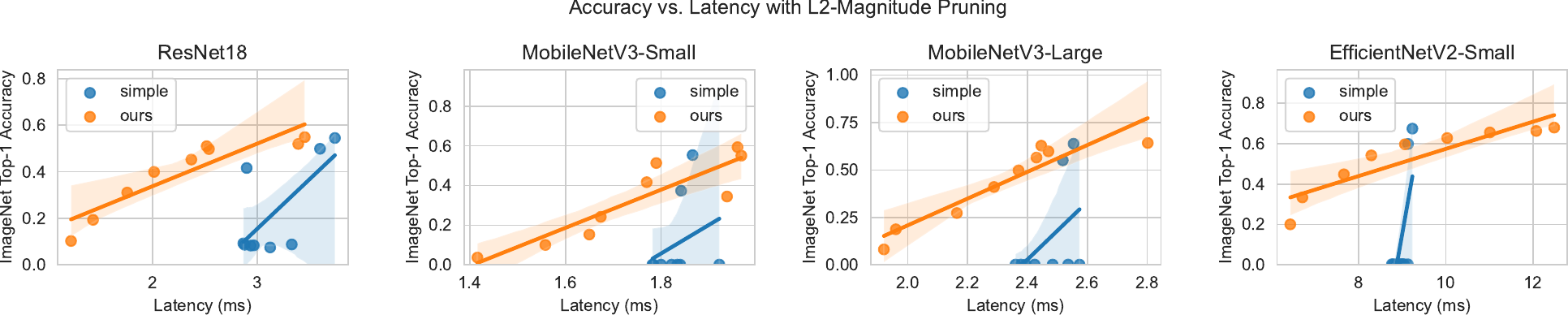}
    \caption{\textbf{Accuracy vs Latency with L2-Magnitude Pruning for Single vs. Multi-branch Segments} This shows tradeoff curves for pruning only single-branch segments of a network (blue), comparing that with pruning all segments -- including multi-branch -- segments of a network (orange). The accuracy-latency tradeoff curve for pruning all segments is more favorable.}
    \label{fig:acc_vs_latency_ft}
\end{figure}

\textbf{Algorithms obtain ``free'' improvements} by pruning multi-branch segments using our export, of up to 14.1\% accuracy points and 29.4\% latency. We assess this improvement by comparing single-branch pruning performance against ``our'' method: (1) Prune all segments in the model, instead of only single-branch segments; (2) Fine-tune for 1 epoch on ImageNet; (3) Evaluate for accuracy, and benchmark both models for latency. This allows us to make an accuracy-latency comparison. We show accuracy-latency tradeoff curves on GPU (Figure \ref{fig:acc_vs_latency_ft}, Table \ref{tab:acc_vs_latency_ft}).

To break down these accuracy-latency tradeoff curves above, we show that (1) accuracy improves when we additionally prune multi-branch segments and (2) latency improves when we use reordering to export multi-branch segments.

\begin{figure}
    \centering
    \includegraphics[width=\textwidth]{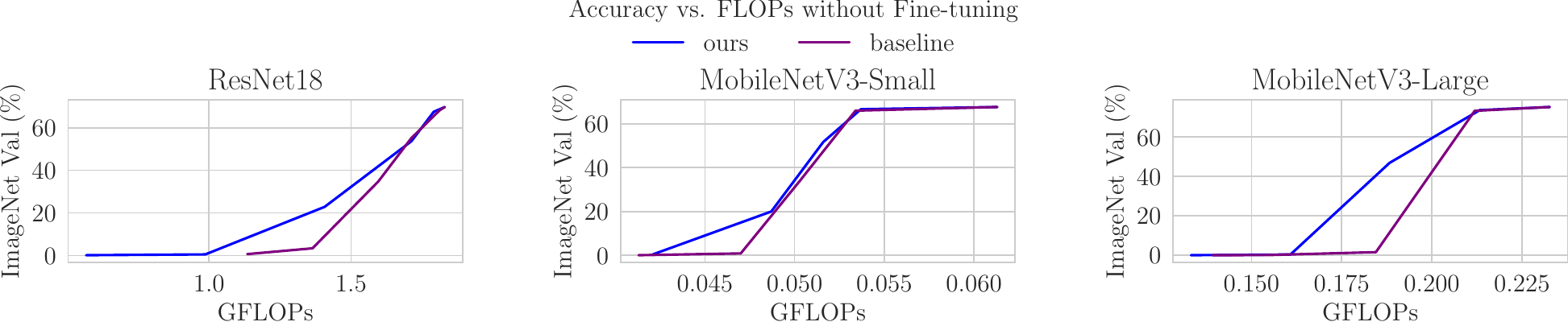}
    \caption{\textbf{Accuracy vs. FLOPs without Fine-Tuning for Single vs. Multi-branch Segments}: Existing pruning algorithms obtain a more favorable accuracy-computation trade off. Note that lower FLOPs (left) and higher accuracy (top) is preferred, so tradeoff curves closer to the top-left are preferred. In particular, pruning all segments attains higher accuracy than pruning single-branch segments only, under the same FLOP budget. Results are obtained using L2-magnitude pruning and without fine-tuning. See Table A.\ref{tab:acc_vs_flops_noft} for numerical results.}
    \label{fig:acc_vs_flops_noft}
\end{figure}

\textbf{Accuracy improves by up to 45\% when \textit{additionally} pruning multi-branch segments}. We assess the impact of additionally pruning multi-branch segments: Prune all segments, including multi-branch ones; run a previously-developed pruning algorithm; fine-tune for 1 epoch; and assess accuracy. We evaluate on a number of modern and baseline neural networks, showing a better accuracy-FLOPs tradeoff curve, by supporting pruning for multi-branch segments. Note that the pruning algorithm itself remains constant; the only change is which part of the network we prune. We find that pruning all segments -- instead of pruning only single-branch segments -- incurs up to a 45-point increase in accuracy or 14$\times$ reduction in FLOPs (
Table A.\ref{tab:acc_vs_flops_ft}), since single-branch segments account for a limited percentage of the total computational cost (Figure \ref{fig:flop_reduction_over_time}). We also report results without fine-tuning, computing accuracy directly on the post-training pruned network (Figure A.\ref{fig:acc_vs_flops_noft}, Table A.\ref{tab:acc_vs_flops_noft}). Above, we report FLOPs to examine the effects of segment choice independently. For completeness, we also analyze tradeoffs using latency, using our reordering algorithm for export (Figure A.\ref{fig:acc_vs_latency_ft}).

Taken together, the ablations above explain the accuracy-latency improvements.

\begin{table*}
\centering
\scriptsize
\begin{tabular}{l|l|ccccc|ccccc}
\toprule
 & & \multicolumn{5}{c}{Accuracy} & \multicolumn{5}{c}{GFLOPs} \\
Model & Pruning & 1\% & 5\% & 10\% & 20\% & 30\% & 1\% & 5\% & 10\% & 20\% & 30\% \\
\midrule
MobileNetV3-Small & Ours (All) & 66.67\% & 51.71\% & 21.98\% & 0.45\% & -- & 0.0537 & 0.0516 & 0.0487 & 0.0421 & --\\
MobileNetV3-Small & Single-branch & 66.01\% & 34.04\% & 0.89\% & 0.08\% & -- & 0.0534 & 0.0503 & 0.047 & 0.0413 & -- \\
MobileNetV3-Large & Ours (All) & 73.70\% & -- & 46.91\% & 0.35\% & 0.017\% & 0.2133 & -- & 0.1883 & 0.1608 & 0.1332 \\
MobileNetV3-Large & Single-branch & 73.34\% & -- & 1.56\% & 0.09\% & 0.010\% & 0.2119 & -- & 0.1845 & 0.1548 & 0.1392 \\
\bottomrule
\end{tabular}
\caption{\textbf{Accuracy vs. FLOPs without Fine-tuning}: Above we compare end-to-end the accuracy-FLOPs impact of our method; this includes (1) pruning all instead of just single-branch segments and (2) using our export strategy that includes reordering weights. The percentages x\% denote what percent of channels we pruned from the total number of channels in the network. However, note that \textit{Ours (All)} and \textit{Single-branch} under the same percentage are not fully comparable, due to the slightly variable FLOPs shown to the right. As a result, we suggest referencing Figure \ref{fig:acc_vs_flops_noft} for a visual tradeoff curve.}
\label{tab:acc_vs_flops_noft}
\end{table*}

\begin{figure}
    \centering
    \includegraphics[width=\textwidth]{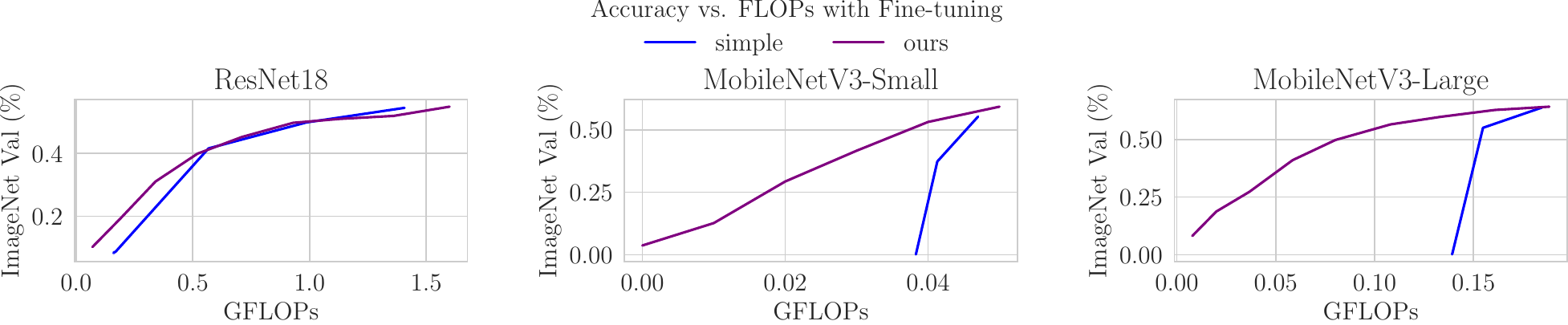}
    \caption{\textbf{Accuracy vs. FLOPs with Fine-Tuning for Single vs. Multi-branch Segments}: Using \Ours, existing pruning algorithms obtain a more favorable accuracy-latency trade off. Note that lower latency (left) and higher accuracy (top) is preferred, so tradeoff curves closer to the top-left are preferred. In particular, pruning all segments attains higher accuracy than pruning single-branch segments only, under the same latency budget. Furthermore, employing our export algorithm ensures that that the final exported, pruned model achieves latency reductions. Results are obtained using L2-magnitude pruning and \textit{with} fine-tuning.}
    \label{fig:acc_vs_flops_ft}
\end{figure}

\section{Unconstrained Input Pruning Accuracy}

\begin{figure}
    \centering
    \includegraphics[width=.95\textwidth]{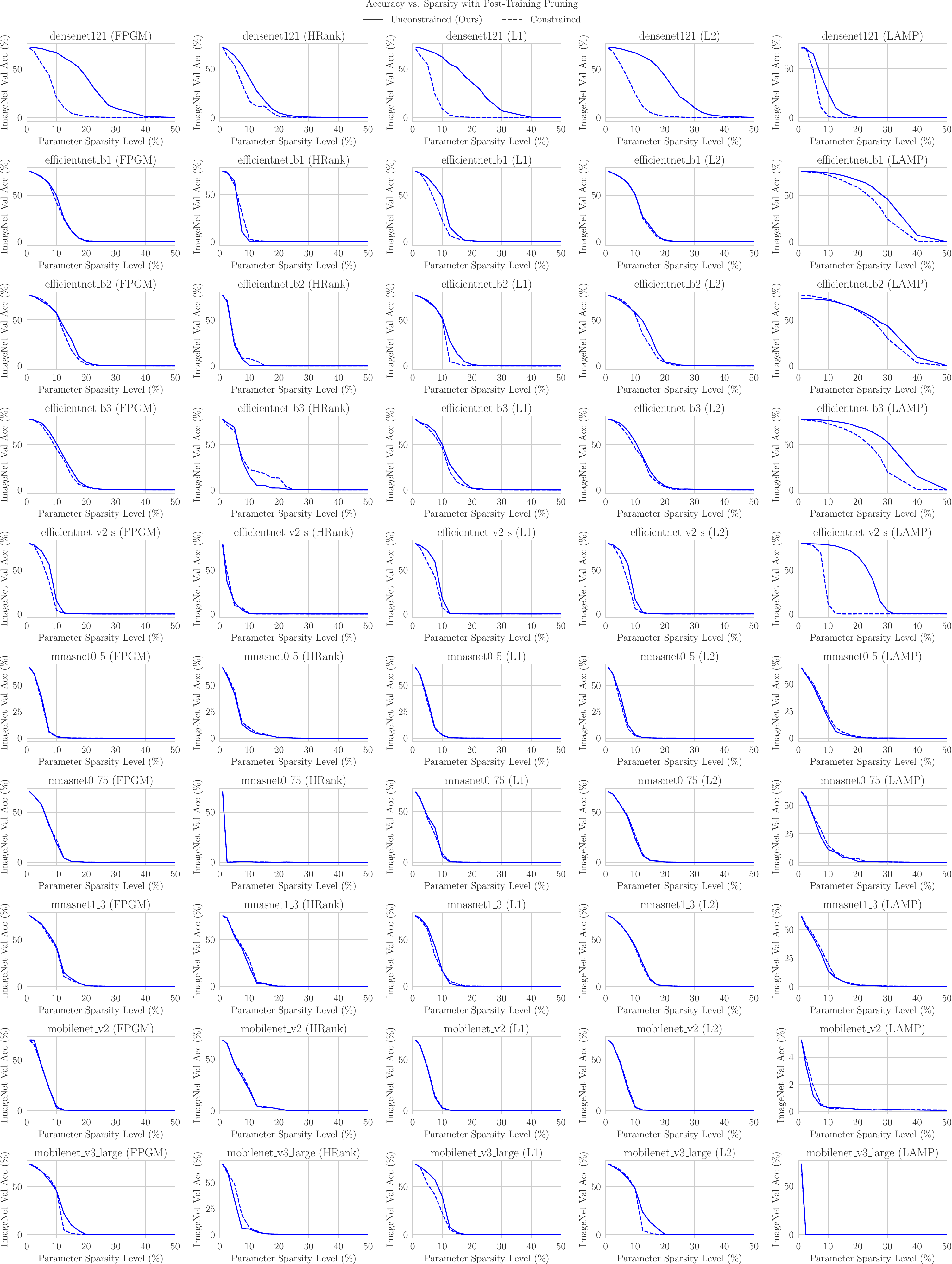}
    \caption{\textbf{Unconstrained vs. constrained accuracy}. In the vast majority of cases, unconstrained accuracy matches or outperforms constrained accuracy. See numeric results for these plots in Tables A.\ref{tab:upscale_vs_naive_latency_0}, A.\ref{tab:upscale_vs_naive_latency_1}, A.\ref{tab:upscale_vs_naive_latency_2}.}
    \label{fig:unconstrained_vs_constrained_accuracy_1}
\end{figure}

\begin{figure*}
    \centering
    \includegraphics[width=\textwidth]{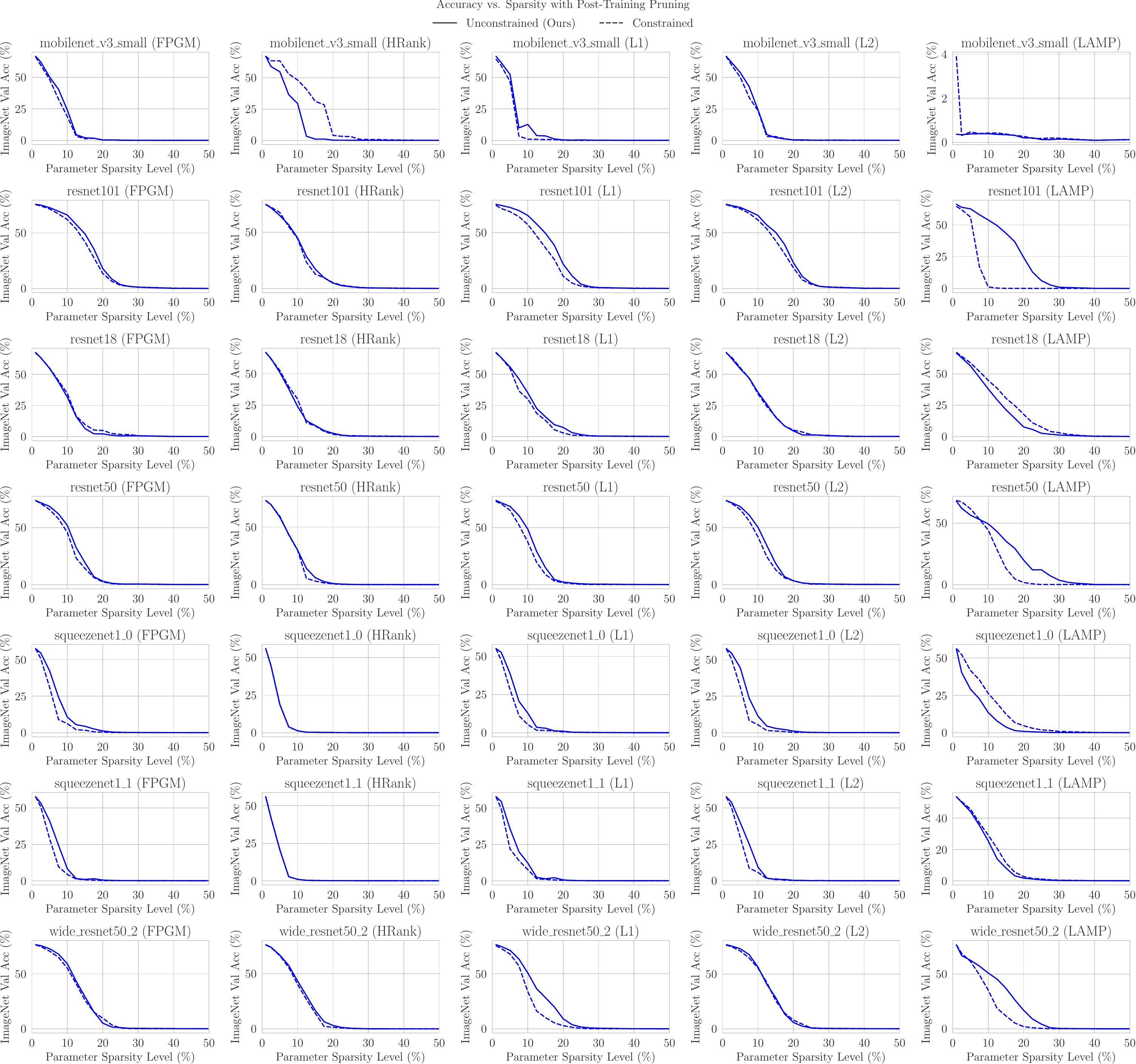}
    \caption{\textbf{Unconstrained vs. constrained accuracy} In the vast majority of cases, unconstrained accuracy matches or outperforms constrained accuracy. See numeric results for these plots in Tables A.\ref{tab:upscale_vs_naive_latency_0}, A.\ref{tab:upscale_vs_naive_latency_1}, A.\ref{tab:upscale_vs_naive_latency_2}.}
    \label{fig:unconstrained_vs_constrained_accuracy_2}
\end{figure*}

This section contains the full set of results for unconstrained input pruning across all models, sparsity levels, and heuristics. These are split between two large grids of figures (Figures \ref{fig:unconstrained_vs_constrained_accuracy_1} and \ref{fig:unconstrained_vs_constrained_accuracy_2}), where we observe that unconstrained accuracy generally matches or outperforms constrained accuracy. We report numerical results in Tables A.\ref{tab:upscale_vs_naive_latency_0}, A.\ref{tab:upscale_vs_naive_latency_1}, A.\ref{tab:upscale_vs_naive_latency_2}.

\section{Unconstrained Input Pruning Latency}

\begin{figure}
    \centering
    \includegraphics[width=.85\textwidth]{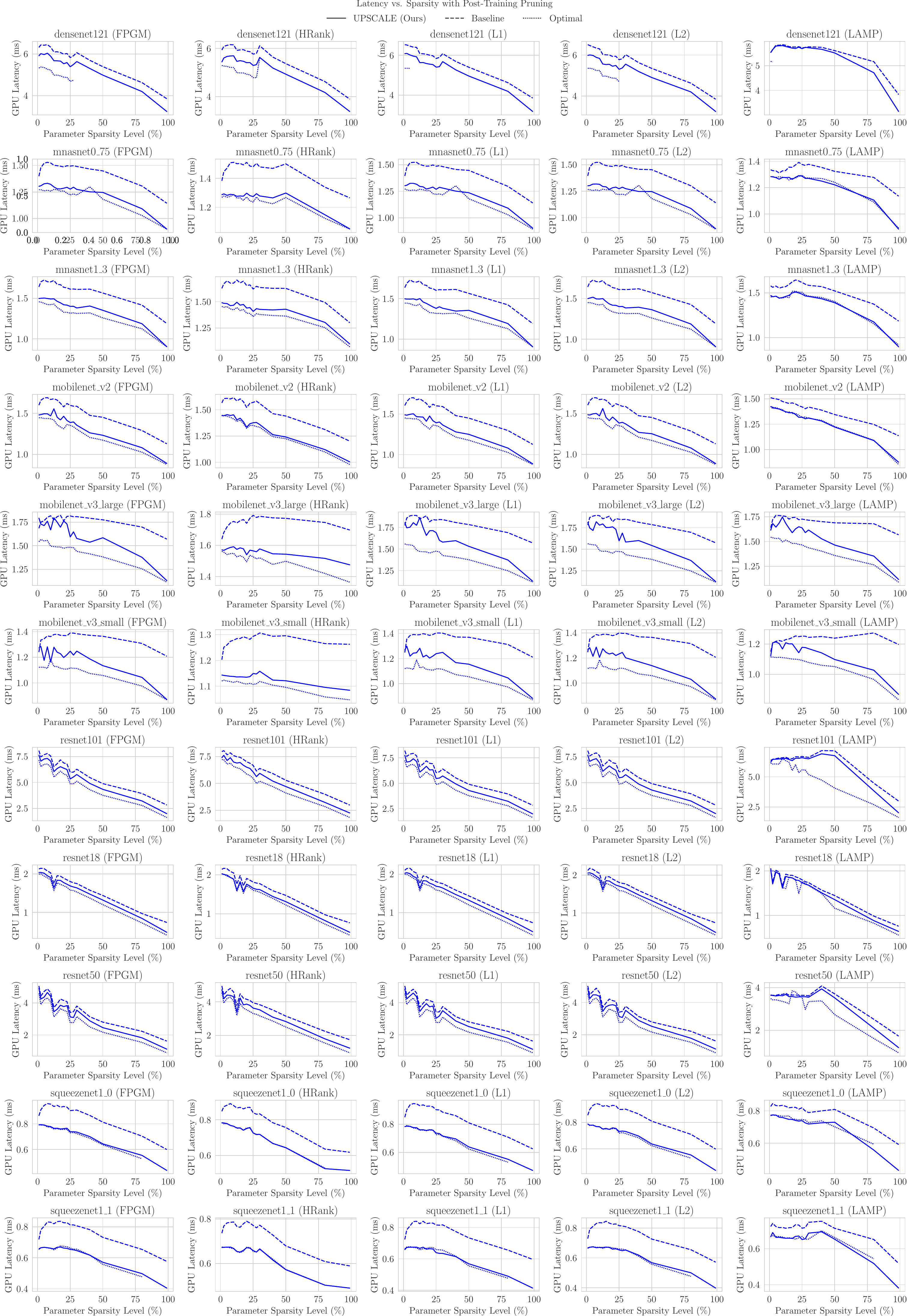}
    \caption{\textbf{Export latency} for our unconstrained export, baseline unconstrained export, and the optimal zero-copy export. Notice that \Ours~(solid) approaches the optimal (dotted) latency fairly often. See numeric results for these plots in Tables A.\ref{tab:upscale_vs_naive_latency_0}, A.\ref{tab:upscale_vs_naive_latency_1}, A.\ref{tab:upscale_vs_naive_latency_2}.}
    \label{fig:unconstrained_vs_constrained_latency}
\end{figure}

This section contains the full set of results for unconstrained input pruning export latency across all models, heuristics, and sparsity levels. We plot the results in A.\ref{fig:unconstrained_vs_constrained_latency} and include the results in table-form in Tables A.\ref{tab:upscale_vs_naive_latency_0}, A.\ref{tab:upscale_vs_naive_latency_1}, A.\ref{tab:upscale_vs_naive_latency_2}.

\begin{table}
    \centering
    \tiny
    \begin{tabular}{l|l|cccccccccc}
    \toprule
    Model & Heuristic & Stat & 1\% & 5\% & 10\% & 15\% & 20\% & 25\% & 30\% & 40\% \\

\midrule
densenet121 
& FPGM & Acc (Ours) & 72.62\% & 71.04\% & 67.03\% & 57.51\% & 42.15\% & 21.57\% & 9.77\% & 1.17\% \\
& & Acc (Cons) & 71.50\% & 55.42\% & 20.77\% & 4.72\% & 1.20\% & 0.47\% & 0.31\% & 0.08\% \\
& & Lat (Ours) & 5.91 ± 0.012 & 5.97 ± 0.006 & 5.89 ± 0.011 & 5.67 ± 0.010 & 5.53 ± 0.007 & 5.39 ± 0.010 & 5.63 ± 0.006 & 5.30 ± 0.003 \\
& & Lat (Base) & 6.31 ± 0.006 & 6.42 ± 0.011 & 6.33 ± 0.012 & 6.10 ± 0.025 & 5.99 ± 0.007 & 5.82 ± 0.018 & 6.06 ± 0.016 & 5.69 ± 0.006 \\
& & Lat (Zero) & 5.33 ± 0.007 & 5.35 ± 0.012 & 5.23 ± 0.007 & 4.99 ± 0.006 & 4.89 ± 0.013 & 4.69 ± 0.004 &  &  \\
\cmidrule{2-11}
& HRank & Acc (Ours) & 72.04\% & 63.59\% & 40.68\% & 17.95\% & 4.87\% & 1.59\% & 0.62\% & 0.17\% \\
& & Acc (Cons) & 72.36\% & 54.10\% & 17.03\% & 11.95\% & 1.46\% & 0.47\% & 0.11\% & 0.11\% \\
& & Lat (Ours) & 5.44 ± 0.018 & 5.67 ± 0.016 & 5.69 ± 0.010 & 5.49 ± 0.005 & 5.41 ± 0.008 & 5.29 ± 0.005 & 5.63 ± 0.008 & 5.19 ± 0.005 \\
& & Lat (Base) & 5.93 ± 0.007 & 6.13 ± 0.004 & 6.15 ± 0.017 & 5.94 ± 0.011 & 5.89 ± 0.013 & 5.76 ± 0.008 & 6.11 ± 0.008 & 5.64 ± 0.010 \\
& & Lat (Zero) & 5.29 ± 0.012 & 5.25 ± 0.003 & 5.21 ± 0.008 & 4.98 ± 0.004 & 4.93 ± 0.005 & 4.79 ± 0.003 & 5.59 ± 0.004 &  \\
\cmidrule{2-11}
& L1 & Acc (Ours) & 72.47\% & 69.16\% & 62.27\% & 51.50\% & 36.07\% & 19.55\% & 7.13\% & 0.43\% \\
& & Acc (Cons) & 70.82\% & 54.89\% & 9.04\% & 1.08\% & 0.38\% & 0.18\% & 0.20\% & 0.11\% \\
& & Lat (Ours) & 6.08 ± 0.009 & 6.02 ± 0.015 & 5.92 ± 0.008 & 5.64 ± 0.013 & 5.56 ± 0.006 & 5.41 ± 0.005 & 5.68 ± 0.009 & 5.27 ± 0.005 \\
& & Lat (Base) & 6.53 ± 0.014 & 6.46 ± 0.004 & 6.40 ± 0.016 & 6.08 ± 0.010 & 6.01 ± 0.013 & 5.84 ± 0.008 & 6.08 ± 0.007 & 5.69 ± 0.007 \\
& & Lat (Zero) & 5.35 ± 0.010 & 5.34 ± 0.010 &  &  &  &  &  &  \\
\cmidrule{2-11}
& L2 & Acc (Ours) & 72.57\% & 70.89\% & 66.64\% & 59.34\% & 43.06\% & 21.59\% & 10.18\% & 1.36\% \\
& & Acc (Cons) & 71.51\% & 55.08\% & 24.76\% & 5.44\% & 1.27\% & 0.64\% & 0.35\% & 0.14\% \\
& & Lat (Ours) & 5.99 ± 0.007 & 5.98 ± 0.009 & 5.84 ± 0.005 & 5.54 ± 0.006 & 5.46 ± 0.010 & 5.31 ± 0.014 & 5.56 ± 0.006 & 5.20 ± 0.002 \\
& & Lat (Base) & 6.51 ± 0.012 & 6.41 ± 0.004 & 6.31 ± 0.012 & 6.01 ± 0.003 & 5.93 ± 0.007 & 5.78 ± 0.009 & 6.02 ± 0.009 & 5.63 ± 0.006 \\
& & Lat (Zero) & 5.36 ± 0.009 & 5.33 ± 0.014 & 5.25 ± 0.002 & 4.98 ± 0.008 & 4.88 ± 0.012 & 4.71 ± 0.010 &  &  \\
\cmidrule{2-11}
& LAMP & Acc (Ours) & 71.80\% & 64.66\% & 25.82\% & 4.33\% & 0.37\% & 0.20\% & 0.11\% & 0.09\% \\
& & Acc (Cons) & 70.64\% & 48.28\% & 1.48\% & 0.12\% & 0.12\% & 0.14\% & 0.15\% & 0.10\% \\
& & Lat (Ours) & 5.53 ± 0.013 & 5.79 ± 0.002 & 5.81 ± 0.007 & 5.74 ± 0.005 & 5.70 ± 0.005 & 5.69 ± 0.012 & 5.72 ± 0.006 & 5.66 ± 0.009 \\
& & Lat (Base) & 5.53 ± 0.005 & 5.83 ± 0.013 & 5.85 ± 0.003 & 5.77 ± 0.009 & 5.75 ± 0.005 & 5.72 ± 0.011 & 5.76 ± 0.009 & 5.74 ± 0.003 \\
& & Lat (Zero) & 5.17 ± 0.010 &  &  &  &  &  &  &  \\
\midrule
mnasnet0\_75 
& FPGM & Acc (Ours) & 70.40\% & 57.28\% & 19.05\% & 1.01\% & 0.17\% & 0.15\% & 0.24\% & 0.10\% \\
& & Acc (Cons) & 70.44\% & 57.24\% & 22.56\% & 1.06\% & 0.25\% & 0.16\% & 0.13\% & 0.07\% \\
& & Lat (Ours) & 1.30 ± 0.001 & 1.33 ± 0.003 & 1.32 ± 0.001 & 1.28 ± 0.006 & 1.29 ± 0.002 & 1.28 ± 0.003 & 1.27 ± 0.004 & 1.25 ± 0.003 \\
& & Lat (Base) & 1.40 ± 0.003 & 1.52 ± 0.003 & 1.52 ± 0.004 & 1.50 ± 0.003 & 1.49 ± 0.001 & 1.50 ± 0.005 & 1.49 ± 0.004 & 1.47 ± 0.004 \\
& & Lat (Zero) & 1.27 ± 0.003 & 1.26 ± 0.002 & 1.26 ± 0.001 & 1.27 ± 0.002 & 1.26 ± 0.002 & 1.22 ± 0.002 & 1.22 ± 0.004 & 1.30 ± 0.006 \\
\cmidrule{2-11}
& HRank & Acc (Ours) & 70.38\% & 0.32\% & 0.55\% & 0.22\% & 0.13\% & 0.16\% & 0.14\% & 0.10\% \\
& & Acc (Cons) & 70.35\% & 0.53\% & 0.86\% & 0.38\% & 0.11\% & 0.12\% & 0.08\% & 0.12\% \\
& & Lat (Ours) & 1.29 ± 0.004 & 1.28 ± 0.003 & 1.29 ± 0.004 & 1.28 ± 0.002 & 1.30 ± 0.002 & 1.27 ± 0.006 & 1.28 ± 0.003 & 1.26 ± 0.005 \\
& & Lat (Base) & 1.38 ± 0.004 & 1.47 ± 0.003 & 1.51 ± 0.004 & 1.50 ± 0.003 & 1.51 ± 0.003 & 1.48 ± 0.005 & 1.48 ± 0.002 & 1.47 ± 0.001 \\
& & Lat (Zero) & 1.27 ± 0.004 & 1.27 ± 0.002 & 1.29 ± 0.004 & 1.26 ± 0.004 & 1.27 ± 0.002 & 1.24 ± 0.002 & 1.25 ± 0.006 & 1.22 ± 0.002 \\
\cmidrule{2-11}
& L1 & Acc (Ours) & 69.47\% & 45.56\% & 5.90\% & 0.35\% & 0.16\% & 0.15\% & 0.16\% & 0.09\% \\
& & Acc (Cons) & 69.69\% & 43.38\% & 8.49\% & 0.43\% & 0.16\% & 0.12\% & 0.13\% & 0.11\% \\
& & Lat (Ours) & 1.30 ± 0.003 & 1.32 ± 0.003 & 1.32 ± 0.006 & 1.29 ± 0.004 & 1.28 ± 0.001 & 1.27 ± 0.004 & 1.28 ± 0.003 & 1.25 ± 0.002 \\
& & Lat (Base) & 1.39 ± 0.003 & 1.52 ± 0.003 & 1.52 ± 0.001 & 1.49 ± 0.002 & 1.49 ± 0.005 & 1.49 ± 0.006 & 1.48 ± 0.004 & 1.46 ± 0.003 \\
& & Lat (Zero) & 1.27 ± 0.002 & 1.26 ± 0.002 & 1.26 ± 0.003 & 1.26 ± 0.006 & 1.26 ± 0.003 & 1.22 ± 0.004 & 1.22 ± 0.002 & 1.30 ± 0.005 \\
\cmidrule{2-11}
& L2 & Acc (Ours) & 70.48\% & 57.40\% & 23.56\% & 1.88\% & 0.36\% & 0.13\% & 0.16\% & 0.10\% \\
& & Acc (Cons) & 70.45\% & 57.46\% & 26.91\% & 2.17\% & 0.32\% & 0.10\% & 0.19\% & 0.09\% \\
& & Lat (Ours) & 1.30 ± 0.007 & 1.32 ± 0.002 & 1.32 ± 0.001 & 1.29 ± 0.001 & 1.28 ± 0.003 & 1.27 ± 0.001 & 1.27 ± 0.002 & 1.25 ± 0.004 \\
& & Lat (Base) & 1.40 ± 0.004 & 1.51 ± 0.004 & 1.52 ± 0.004 & 1.49 ± 0.003 & 1.48 ± 0.000 & 1.49 ± 0.003 & 1.48 ± 0.005 & 1.46 ± 0.001 \\
& & Lat (Zero) & 1.27 ± 0.002 & 1.26 ± 0.003 & 1.26 ± 0.004 & 1.26 ± 0.002 & 1.26 ± 0.002 & 1.22 ± 0.002 & 1.22 ± 0.001 & 1.31 ± 0.003 \\
\cmidrule{2-11}
& LAMP & Acc (Ours) & 61.90\% & 40.28\% & 11.29\% & 4.31\% & 0.93\% & 0.68\% & 0.45\% & 0.10\% \\
& & Acc (Cons) & 62.01\% & 41.21\% & 14.68\% & 5.82\% & 3.31\% & 0.79\% & 0.39\% & 0.11\% \\
& & Lat (Ours) & 1.28 ± 0.002 & 1.28 ± 0.002 & 1.28 ± 0.003 & 1.28 ± 0.006 & 1.28 ± 0.003 & 1.29 ± 0.006 & 1.27 ± 0.001 & 1.25 ± 0.003 \\
& & Lat (Base) & 1.34 ± 0.003 & 1.33 ± 0.004 & 1.33 ± 0.002 & 1.36 ± 0.002 & 1.37 ± 0.002 & 1.38 ± 0.002 & 1.38 ± 0.002 & 1.35 ± 0.010 \\
& & Lat (Zero) & 1.28 ± 0.004 & 1.28 ± 0.003 & 1.28 ± 0.003 & 1.26 ± 0.002 & 1.27 ± 0.002 & 1.29 ± 0.005 & 1.27 ± 0.002 & 1.27 ± 0.003 \\
\midrule
mnasnet1\_3 
& FPGM & Acc (Ours) & 75.48\% & 66.67\% & 42.70\% & 8.37\% & 0.79\% & 0.33\% & 0.21\% & 0.17\% \\
& & Acc (Cons) & 75.47\% & 65.90\% & 41.05\% & 6.37\% & 0.87\% & 0.31\% & 0.15\% & 0.10\% \\
& & Lat (Ours) & 1.50 ± 0.005 & 1.50 ± 0.003 & 1.49 ± 0.003 & 1.46 ± 0.002 & 1.42 ± 0.002 & 1.40 ± 0.003 & 1.38 ± 0.002 & 1.41 ± 0.004 \\
& & Lat (Base) & 1.64 ± 0.002 & 1.73 ± 0.004 & 1.70 ± 0.002 & 1.67 ± 0.006 & 1.64 ± 0.003 & 1.61 ± 0.008 & 1.61 ± 0.001 & 1.61 ± 0.002 \\
& & Lat (Zero) & 1.46 ± 0.002 & 1.44 ± 0.004 & 1.42 ± 0.004 & 1.39 ± 0.003 & 1.33 ± 0.004 & 1.32 ± 0.003 & 1.31 ± 0.003 & 1.32 ± 0.006 \\
\cmidrule{2-11}
& HRank & Acc (Ours) & 75.24\% & 53.23\% & 20.99\% & 2.95\% & 0.24\% & 0.08\% & 0.10\% & 0.10\% \\
& & Acc (Cons) & 75.39\% & 55.13\% & 28.11\% & 3.48\% & 0.23\% & 0.14\% & 0.09\% & 0.13\% \\
& & Lat (Ours) & 1.49 ± 0.001 & 1.48 ± 0.003 & 1.47 ± 0.005 & 1.47 ± 0.002 & 1.44 ± 0.002 & 1.41 ± 0.006 & 1.43 ± 0.004 & 1.42 ± 0.003 \\
& & Lat (Base) & 1.63 ± 0.002 & 1.70 ± 0.002 & 1.68 ± 0.004 & 1.68 ± 0.003 & 1.65 ± 0.003 & 1.62 ± 0.003 & 1.64 ± 0.004 & 1.63 ± 0.004 \\
& & Lat (Zero) & 1.45 ± 0.003 & 1.45 ± 0.006 & 1.42 ± 0.002 & 1.42 ± 0.004 & 1.39 ± 0.002 & 1.36 ± 0.004 & 1.38 ± 0.006 & 1.37 ± 0.004 \\
\cmidrule{2-11}
& L1 & Acc (Ours) & 75.50\% & 63.88\% & 16.89\% & 0.99\% & 0.10\% & 0.10\% & 0.09\% & 0.10\% \\
& & Acc (Cons) & 75.11\% & 61.19\% & 16.30\% & 2.85\% & 0.32\% & 0.13\% & 0.11\% & 0.10\% \\
& & Lat (Ours) & 1.50 ± 0.004 & 1.49 ± 0.003 & 1.49 ± 0.002 & 1.46 ± 0.001 & 1.42 ± 0.003 & 1.38 ± 0.003 & 1.38 ± 0.003 & 1.35 ± 0.004 \\
& & Lat (Base) & 1.64 ± 0.001 & 1.73 ± 0.003 & 1.71 ± 0.003 & 1.67 ± 0.003 & 1.63 ± 0.002 & 1.60 ± 0.003 & 1.61 ± 0.005 & 1.61 ± 0.003 \\
& & Lat (Zero) & 1.44 ± 0.003 & 1.44 ± 0.001 & 1.42 ± 0.001 & 1.38 ± 0.003 & 1.33 ± 0.004 & 1.32 ± 0.003 & 1.31 ± 0.003 & 1.32 ± 0.003 \\
\cmidrule{2-11}
& L2 & Acc (Ours) & 75.41\% & 66.44\% & 43.26\% & 7.88\% & 0.78\% & 0.16\% & 0.10\% & 0.11\% \\
& & Acc (Cons) & 75.42\% & 65.88\% & 41.38\% & 6.92\% & 0.73\% & 0.22\% & 0.20\% & 0.10\% \\
& & Lat (Ours) & 1.50 ± 0.001 & 1.52 ± 0.001 & 1.49 ± 0.006 & 1.46 ± 0.003 & 1.42 ± 0.003 & 1.40 ± 0.002 & 1.38 ± 0.003 & 1.39 ± 0.001 \\
& & Lat (Base) & 1.64 ± 0.001 & 1.73 ± 0.003 & 1.70 ± 0.005 & 1.67 ± 0.003 & 1.63 ± 0.004 & 1.61 ± 0.004 & 1.61 ± 0.001 & 1.62 ± 0.006 \\
& & Lat (Zero) & 1.45 ± 0.005 & 1.44 ± 0.002 & 1.42 ± 0.002 & 1.38 ± 0.002 & 1.33 ± 0.004 & 1.32 ± 0.004 & 1.31 ± 0.004 & 1.32 ± 0.003 \\
\cmidrule{2-11}
& LAMP & Acc (Ours) & 61.34\% & 42.75\% & 13.82\% & 4.54\% & 1.11\% & 0.60\% & 0.14\% & 0.12\% \\
& & Acc (Cons) & 62.08\% & 45.27\% & 19.84\% & 4.38\% & 1.40\% & 0.82\% & 0.31\% & 0.13\% \\
& & Lat (Ours) & 1.47 ± 0.004 & 1.46 ± 0.004 & 1.45 ± 0.003 & 1.47 ± 0.002 & 1.51 ± 0.003 & 1.49 ± 0.004 & 1.46 ± 0.003 & 1.43 ± 0.003 \\
& & Lat (Base) & 1.58 ± 0.005 & 1.58 ± 0.004 & 1.56 ± 0.005 & 1.59 ± 0.002 & 1.64 ± 0.001 & 1.62 ± 0.001 & 1.58 ± 0.003 & 1.57 ± 0.003 \\
& & Lat (Zero) & 1.45 ± 0.003 & 1.46 ± 0.002 & 1.44 ± 0.004 & 1.45 ± 0.004 & 1.52 ± 0.003 & 1.51 ± 0.002 & 1.46 ± 0.003 & 1.45 ± 0.002 \\

        \bottomrule
        \end{tabular}
        \caption{ \textbf{UPSCALE accuracy and latency} across architectures, sparsity levels, and heuristics. Notice that latency for ours is comparable to the ideal, zero-copy reference, much lower than the baseline export's latency. Additionally notice our (unconstrained) accuracy matches or outperforms the baseline constrained accuracy. }
        \label{tab:upscale_vs_naive_latency_0}
    \end{table}\begin{table}
    \centering
    \tiny
    \begin{tabular}{l|l|cccccccccc}
    \toprule
    Model & Heuristic & Stat & 1\% & 5\% & 10\% & 15\% & 20\% & 25\% & 30\% & 40\% \\

\midrule
mobilenet\_v2 
& FPGM & Acc (Ours) & 69.99\% & 43.69\% & 2.82\% & 0.28\% & 0.13\% & 0.10\% & 0.09\% & 0.10\% \\
& & Acc (Cons) & 69.54\% & 44.50\% & 4.18\% & 0.38\% & 0.14\% & 0.10\% & 0.10\% & 0.13\% \\
& & Lat (Ours) & 1.48 ± 0.002 & 1.49 ± 0.006 & 1.47 ± 0.005 & 1.45 ± 0.002 & 1.46 ± 0.005 & 1.39 ± 0.007 & 1.36 ± 0.004 & 1.26 ± 0.004 \\
& & Lat (Base) & 1.60 ± 0.002 & 1.69 ± 0.004 & 1.67 ± 0.005 & 1.66 ± 0.002 & 1.58 ± 0.002 & 1.60 ± 0.007 & 1.59 ± 0.006 & 1.48 ± 0.005 \\
& & Lat (Zero) & 1.45 ± 0.003 & 1.44 ± 0.005 & 1.43 ± 0.004 & 1.36 ± 0.004 & 1.31 ± 0.007 & 1.35 ± 0.004 & 1.31 ± 0.005 & 1.21 ± 0.008 \\
\cmidrule{2-11}
& HRank & Acc (Ours) & 68.52\% & 45.15\% & 19.83\% & 3.00\% & 1.53\% & 0.21\% & 0.15\% & 0.11\% \\
& & Acc (Cons) & 68.51\% & 45.71\% & 21.60\% & 3.64\% & 1.68\% & 0.26\% & 0.20\% & 0.10\% \\
& & Lat (Ours) & 1.44 ± 0.008 & 1.45 ± 0.002 & 1.45 ± 0.003 & 1.42 ± 0.003 & 1.33 ± 0.004 & 1.38 ± 0.005 & 1.36 ± 0.003 & 1.26 ± 0.001 \\
& & Lat (Base) & 1.57 ± 0.004 & 1.61 ± 0.003 & 1.62 ± 0.002 & 1.60 ± 0.002 & 1.52 ± 0.005 & 1.57 ± 0.003 & 1.56 ± 0.004 & 1.46 ± 0.002 \\
& & Lat (Zero) & 1.45 ± 0.002 & 1.44 ± 0.006 & 1.41 ± 0.004 & 1.40 ± 0.004 & 1.32 ± 0.002 & 1.35 ± 0.002 & 1.33 ± 0.002 & 1.24 ± 0.004 \\
\cmidrule{2-11}
& L1 & Acc (Ours) & 69.35\% & 41.83\% & 2.37\% & 0.23\% & 0.13\% & 0.11\% & 0.11\% & 0.10\% \\
& & Acc (Cons) & 69.35\% & 42.92\% & 2.81\% & 0.16\% & 0.12\% & 0.13\% & 0.10\% & 0.16\% \\
& & Lat (Ours) & 1.49 ± 0.002 & 1.50 ± 0.005 & 1.47 ± 0.005 & 1.46 ± 0.005 & 1.48 ± 0.007 & 1.39 ± 0.004 & 1.36 ± 0.003 & 1.28 ± 0.003 \\
& & Lat (Base) & 1.61 ± 0.004 & 1.70 ± 0.003 & 1.68 ± 0.003 & 1.66 ± 0.002 & 1.58 ± 0.004 & 1.60 ± 0.002 & 1.59 ± 0.006 & 1.48 ± 0.006 \\
& & Lat (Zero) & 1.45 ± 0.005 & 1.45 ± 0.004 & 1.44 ± 0.004 & 1.37 ± 0.004 & 1.31 ± 0.004 & 1.36 ± 0.011 & 1.31 ± 0.012 & 1.22 ± 0.007 \\
\cmidrule{2-11}
& L2 & Acc (Ours) & 69.48\% & 46.52\% & 3.08\% & 0.27\% & 0.18\% & 0.10\% & 0.10\% & 0.10\% \\
& & Acc (Cons) & 69.60\% & 47.78\% & 3.89\% & 0.39\% & 0.14\% & 0.10\% & 0.10\% & 0.10\% \\
& & Lat (Ours) & 1.48 ± 0.004 & 1.50 ± 0.005 & 1.47 ± 0.007 & 1.45 ± 0.006 & 1.46 ± 0.005 & 1.38 ± 0.002 & 1.37 ± 0.002 & 1.27 ± 0.004 \\
& & Lat (Base) & 1.60 ± 0.005 & 1.69 ± 0.005 & 1.67 ± 0.004 & 1.66 ± 0.004 & 1.58 ± 0.008 & 1.60 ± 0.006 & 1.58 ± 0.004 & 1.47 ± 0.003 \\
& & Lat (Zero) & 1.45 ± 0.003 & 1.44 ± 0.005 & 1.44 ± 0.004 & 1.37 ± 0.006 & 1.31 ± 0.002 & 1.35 ± 0.003 & 1.30 ± 0.003 & 1.21 ± 0.002 \\
\cmidrule{2-11}
& LAMP & Acc (Ours) & 5.28\% & 1.15\% & 0.29\% & 0.24\% & 0.15\% & 0.10\% & 0.14\% & 0.10\% \\
& & Acc (Cons) & 5.24\% & 1.87\% & 0.27\% & 0.26\% & 0.18\% & 0.13\% & 0.10\% & 0.13\% \\
& & Lat (Ours) & 1.42 ± 0.006 & 1.41 ± 0.004 & 1.39 ± 0.004 & 1.36 ± 0.003 & 1.35 ± 0.001 & 1.32 ± 0.004 & 1.31 ± 0.001 & 1.28 ± 0.002 \\
& & Lat (Base) & 1.51 ± 0.005 & 1.50 ± 0.001 & 1.47 ± 0.003 & 1.45 ± 0.004 & 1.45 ± 0.004 & 1.42 ± 0.003 & 1.42 ± 0.001 & 1.39 ± 0.004 \\
& & Lat (Zero) & 1.43 ± 0.004 & 1.41 ± 0.003 & 1.39 ± 0.004 & 1.37 ± 0.005 & 1.36 ± 0.003 & 1.32 ± 0.004 & 1.31 ± 0.004 & 1.29 ± 0.006 \\
\midrule
mobilenet\_v3\_large 
& FPGM & Acc (Ours) & 73.67\% & 66.05\% & 45.97\% & 10.13\% & 0.29\% & 0.20\% & 0.22\% & 0.10\% \\
& & Acc (Cons) & 73.64\% & 66.13\% & 47.13\% & 1.20\% & 0.21\% & 0.15\% & 0.16\% & 0.13\% \\
& & Lat (Ours) & 1.79 ± 0.002 & 1.71 ± 0.001 & 1.67 ± 0.002 & 1.75 ± 0.005 & 1.76 ± 0.003 & 1.59 ± 0.003 & 1.58 ± 0.005 & 1.54 ± 0.005 \\
& & Lat (Base) & 1.68 ± 0.003 & 1.77 ± 0.003 & 1.82 ± 0.009 & 1.81 ± 0.007 & 1.76 ± 0.007 & 1.82 ± 0.004 & 1.81 ± 0.006 & 1.79 ± 0.004 \\
& & Lat (Zero) & 1.54 ± 0.002 & 1.55 ± 0.003 & 1.49 ± 0.001 & 1.49 ± 0.004 & 1.47 ± 0.006 & 1.48 ± 0.003 & 1.44 ± 0.002 & 1.41 ± 0.004 \\
\cmidrule{2-11}
& HRank & Acc (Ours) & 68.11\% & 33.63\% & 5.46\% & 0.96\% & 0.53\% & 0.22\% & 0.16\% & 0.10\% \\
& & Acc (Cons) & 68.03\% & 49.05\% & 6.72\% & 1.02\% & 0.30\% & 0.24\% & 0.21\% & 0.14\% \\
& & Lat (Ours) & 1.57 ± 0.003 & 1.58 ± 0.003 & 1.59 ± 0.005 & 1.58 ± 0.002 & 1.54 ± 0.002 & 1.56 ± 0.004 & 1.58 ± 0.005 & 1.54 ± 0.003 \\
& & Lat (Base) & 1.64 ± 0.006 & 1.72 ± 0.005 & 1.75 ± 0.006 & 1.76 ± 0.005 & 1.74 ± 0.003 & 1.79 ± 0.006 & 1.79 ± 0.006 & 1.78 ± 0.005 \\
& & Lat (Zero) & 1.56 ± 0.003 & 1.54 ± 0.004 & 1.55 ± 0.006 & 1.53 ± 0.003 & 1.49 ± 0.002 & 1.53 ± 0.004 & 1.52 ± 0.004 & 1.48 ± 0.003 \\
\cmidrule{2-11}
& L1 & Acc (Ours) & 73.72\% & 64.88\% & 39.87\% & 2.23\% & 0.49\% & 0.16\% & 0.21\% & 0.14\% \\
& & Acc (Cons) & 73.69\% & 52.93\% & 22.81\% & 1.06\% & 0.14\% & 0.18\% & 0.15\% & 0.15\% \\
& & Lat (Ours) & 1.80 ± 0.005 & 1.75 ± 0.004 & 1.78 ± 0.005 & 1.75 ± 0.001 & 1.71 ± 0.004 & 1.69 ± 0.001 & 1.58 ± 0.003 & 1.60 ± 0.004 \\
& & Lat (Base) & 1.77 ± 0.004 & 1.88 ± 0.008 & 1.89 ± 0.005 & 1.86 ± 0.003 & 1.82 ± 0.004 & 1.84 ± 0.003 & 1.84 ± 0.003 & 1.81 ± 0.006 \\
& & Lat (Zero) & 1.56 ± 0.001 & 1.55 ± 0.002 & 1.49 ± 0.006 & 1.47 ± 0.002 & 1.45 ± 0.002 & 1.48 ± 0.003 & 1.43 ± 0.003 & 1.40 ± 0.004 \\
\cmidrule{2-11}
& L2 & Acc (Ours) & 73.67\% & 66.30\% & 47.81\% & 13.34\% & 0.41\% & 0.20\% & 0.13\% & 0.14\% \\
& & Acc (Cons) & 73.59\% & 67.34\% & 48.24\% & 1.81\% & 0.24\% & 0.21\% & 0.16\% & 0.14\% \\
& & Lat (Ours) & 1.81 ± 0.005 & 1.72 ± 0.002 & 1.78 ± 0.005 & 1.75 ± 0.006 & 1.76 ± 0.003 & 1.60 ± 0.005 & 1.58 ± 0.005 & 1.60 ± 0.006 \\
& & Lat (Base) & 1.78 ± 0.002 & 1.89 ± 0.004 & 1.89 ± 0.004 & 1.87 ± 0.004 & 1.82 ± 0.002 & 1.85 ± 0.003 & 1.85 ± 0.003 & 1.81 ± 0.008 \\
& & Lat (Zero) & 1.56 ± 0.002 & 1.55 ± 0.004 & 1.49 ± 0.003 & 1.48 ± 0.005 & 1.45 ± 0.002 & 1.47 ± 0.002 & 1.44 ± 0.003 & 1.41 ± 0.003 \\
\cmidrule{2-11}
& LAMP & Acc (Ours) & 72.49\% & 0.08\% & 0.09\% & 0.17\% & 0.18\% & 0.10\% & 0.10\% & 0.10\% \\
& & Acc (Cons) & 68.16\% & 0.11\% & 0.06\% & 0.10\% & 0.11\% & 0.12\% & 0.12\% & 0.08\% \\
& & Lat (Ours) & 1.61 ± 0.004 & 1.65 ± 0.003 & 1.75 ± 0.002 & 1.63 ± 0.002 & 1.63 ± 0.004 & 1.64 ± 0.004 & 1.62 ± 0.003 & 1.53 ± 0.003 \\
& & Lat (Base) & 1.64 ± 0.003 & 1.77 ± 0.002 & 1.76 ± 0.007 & 1.73 ± 0.006 & 1.75 ± 0.008 & 1.73 ± 0.005 & 1.72 ± 0.005 & 1.70 ± 0.012 \\
& & Lat (Zero) & 1.54 ± 0.003 & 1.53 ± 0.003 & 1.50 ± 0.002 & 1.50 ± 0.004 & 1.48 ± 0.001 & 1.47 ± 0.004 & 1.44 ± 0.001 & 1.40 ± 0.004 \\
\midrule
mobilenet\_v3\_small 
& FPGM & Acc (Ours) & 66.74\% & 50.26\% & 24.04\% & 2.09\% & 0.41\% & 0.19\% & 0.11\% & 0.12\% \\
& & Acc (Cons) & 66.45\% & 48.29\% & 18.80\% & 1.37\% & 0.37\% & 0.23\% & 0.14\% & 0.09\% \\
& & Lat (Ours) & 1.27 ± 0.005 & 1.18 ± 0.003 & 1.16 ± 0.002 & 1.24 ± 0.004 & 1.20 ± 0.001 & 1.25 ± 0.005 & 1.25 ± 0.003 & 1.19 ± 0.002 \\
& & Lat (Base) & 1.24 ± 0.003 & 1.34 ± 0.002 & 1.36 ± 0.001 & 1.38 ± 0.003 & 1.37 ± 0.004 & 1.39 ± 0.005 & 1.38 ± 0.002 & 1.37 ± 0.001 \\
& & Lat (Zero) & 1.12 ± 0.004 & 1.12 ± 0.000 & 1.18 ± 0.004 & 1.13 ± 0.002 & 1.11 ± 0.003 & 1.12 ± 0.002 & 1.11 ± 0.001 & 1.07 ± 0.003 \\
\cmidrule{2-11}
& HRank & Acc (Ours) & 67.23\% & 54.81\% & 29.53\% & 1.11\% & 0.26\% & 0.19\% & 0.15\% & 0.09\% \\
& & Acc (Cons) & 67.20\% & 63.55\% & 48.31\% & 31.20\% & 4.08\% & 3.11\% & 0.80\% & 0.30\% \\
& & Lat (Ours) & 1.14 ± 0.005 & 1.14 ± 0.003 & 1.14 ± 0.001 & 1.14 ± 0.002 & 1.14 ± 0.002 & 1.15 ± 0.001 & 1.16 ± 0.003 & 1.12 ± 0.002 \\
& & Lat (Base) & 1.20 ± 0.004 & 1.26 ± 0.001 & 1.28 ± 0.003 & 1.28 ± 0.002 & 1.29 ± 0.006 & 1.29 ± 0.004 & 1.31 ± 0.004 & 1.29 ± 0.004 \\
& & Lat (Zero) & 1.12 ± 0.002 & 1.12 ± 0.003 & 1.11 ± 0.003 & 1.12 ± 0.002 & 1.11 ± 0.003 & 1.11 ± 0.004 & 1.12 ± 0.003 & 1.10 ± 0.002 \\
\cmidrule{2-11}
& L1 & Acc (Ours) & 66.74\% & 52.21\% & 12.74\% & 3.60\% & 0.43\% & 0.44\% & 0.21\% & 0.11\% \\
& & Acc (Cons) & 64.43\% & 46.59\% & 1.02\% & 0.75\% & 0.26\% & 0.12\% & 0.15\% & 0.11\% \\
& & Lat (Ours) & 1.27 ± 0.003 & 1.22 ± 0.006 & 1.25 ± 0.002 & 1.21 ± 0.003 & 1.20 ± 0.001 & 1.24 ± 0.002 & 1.25 ± 0.002 & 1.17 ± 0.002 \\
& & Lat (Base) & 1.25 ± 0.006 & 1.38 ± 0.003 & 1.39 ± 0.004 & 1.40 ± 0.006 & 1.38 ± 0.004 & 1.40 ± 0.003 & 1.40 ± 0.003 & 1.37 ± 0.004 \\
& & Lat (Zero) & 1.12 ± 0.004 & 1.12 ± 0.001 & 1.19 ± 0.007 & 1.13 ± 0.007 & 1.11 ± 0.007 & 1.12 ± 0.005 & 1.11 ± 0.002 & 1.07 ± 0.003 \\
\cmidrule{2-11}
& L2 & Acc (Ours) & 66.86\% & 53.90\% & 24.34\% & 2.21\% & 0.43\% & 0.21\% & 0.19\% & 0.17\% \\
& & Acc (Cons) & 66.89\% & 50.20\% & 23.69\% & 3.00\% & 0.28\% & 0.20\% & 0.25\% & 0.05\% \\
& & Lat (Ours) & 1.27 ± 0.004 & 1.21 ± 0.003 & 1.21 ± 0.002 & 1.25 ± 0.004 & 1.24 ± 0.003 & 1.23 ± 0.002 & 1.20 ± 0.003 & 1.17 ± 0.002 \\
& & Lat (Base) & 1.25 ± 0.003 & 1.37 ± 0.002 & 1.38 ± 0.003 & 1.39 ± 0.003 & 1.38 ± 0.004 & 1.40 ± 0.003 & 1.40 ± 0.005 & 1.37 ± 0.002 \\
& & Lat (Zero) & 1.12 ± 0.002 & 1.12 ± 0.001 & 1.19 ± 0.002 & 1.13 ± 0.005 & 1.11 ± 0.002 & 1.12 ± 0.003 & 1.11 ± 0.002 & 1.07 ± 0.002 \\
\cmidrule{2-11}
& LAMP & Acc (Ours) & 0.37\% & 0.38\% & 0.39\% & 0.35\% & 0.20\% & 0.13\% & 0.15\% & 0.09\% \\
& & Acc (Cons) & 3.91\% & 0.47\% & 0.42\% & 0.38\% & 0.27\% & 0.17\% & 0.19\% & 0.09\% \\
& & Lat (Ours) & 1.12 ± 0.002 & 1.21 ± 0.001 & 1.17 ± 0.004 & 1.20 ± 0.002 & 1.17 ± 0.002 & 1.18 ± 0.004 & 1.17 ± 0.004 & 1.14 ± 0.003 \\
& & Lat (Base) & 1.15 ± 0.002 & 1.22 ± 0.002 & 1.22 ± 0.003 & 1.23 ± 0.003 & 1.25 ± 0.001 & 1.25 ± 0.003 & 1.24 ± 0.004 & 1.25 ± 0.002 \\
& & Lat (Zero) & 1.11 ± 0.003 & 1.11 ± 0.002 & 1.11 ± 0.003 & 1.10 ± 0.002 & 1.10 ± 0.004 & 1.09 ± 0.002 & 1.08 ± 0.003 & 1.06 ± 0.002 \\

        \bottomrule
        \end{tabular}
        \caption{ \textbf{UPSCALE accuracy and latency} across architectures, sparsity levels, and heuristics. Notice that latency for ours is comparable to the ideal, zero-copy reference, much lower than the baseline export's latency. Additionally notice our (unconstrained) accuracy matches or outperforms the baseline constrained accuracy. }
        \label{tab:upscale_vs_naive_latency_1}
    \end{table}\begin{table}
    \centering
    \tiny
    \begin{tabular}{l|l|cccccccccc}
    \toprule
    Model & Heuristic & Stat & 1\% & 5\% & 10\% & 15\% & 20\% & 25\% & 30\% & 40\% \\

\midrule
resnet101 
& FPGM & Acc (Ours) & 75.60\% & 72.61\% & 65.96\% & 49.01\% & 17.87\% & 3.47\% & 1.20\% & 0.20\% \\
& & Acc (Cons) & 75.39\% & 71.41\% & 61.86\% & 41.75\% & 13.50\% & 2.89\% & 1.30\% & 0.32\% \\
& & Lat (Ours) & 7.65 ± 0.009 & 7.27 ± 0.018 & 7.06 ± 0.016 & 6.35 ± 0.008 & 6.38 ± 0.004 & 5.35 ± 0.008 & 5.77 ± 0.007 & 4.92 ± 0.002 \\
& & Lat (Base) & 8.08 ± 0.005 & 7.69 ± 0.023 & 7.56 ± 0.007 & 6.82 ± 0.006 & 6.89 ± 0.006 & 5.99 ± 0.005 & 6.29 ± 0.015 & 5.48 ± 0.007 \\
& & Lat (Zero) & 7.17 ± 0.011 & 6.76 ± 0.012 & 6.53 ± 0.014 & 5.81 ± 0.003 & 5.83 ± 0.008 & 4.87 ± 0.003 & 5.13 ± 0.008 & 4.31 ± 0.005 \\
\cmidrule{2-11}
& HRank & Acc (Ours) & 74.62\% & 64.80\% & 44.93\% & 17.30\% & 4.78\% & 1.58\% & 0.59\% & 0.22\% \\
& & Acc (Cons) & 74.80\% & 67.51\% & 44.38\% & 12.78\% & 5.19\% & 1.85\% & 0.61\% & 0.22\% \\
& & Lat (Ours) & 7.39 ± 0.004 & 7.13 ± 0.008 & 6.99 ± 0.005 & 6.94 ± 0.004 & 6.51 ± 0.012 & 6.06 ± 0.013 & 5.94 ± 0.006 & 5.21 ± 0.014 \\
& & Lat (Base) & 7.96 ± 0.013 & 7.66 ± 0.007 & 7.59 ± 0.015 & 7.43 ± 0.004 & 7.01 ± 0.004 & 6.67 ± 0.011 & 6.39 ± 0.005 & 5.92 ± 0.008 \\
& & Lat (Zero) & 7.22 ± 0.025 & 6.80 ± 0.008 & 6.59 ± 0.008 & 6.37 ± 0.008 & 5.87 ± 0.012 & 5.48 ± 0.007 & 5.20 ± 0.001 & 4.70 ± 0.005 \\
\cmidrule{2-11}
& L1 & Acc (Ours) & 75.44\% & 72.61\% & 65.45\% & 49.48\% & 21.84\% & 3.95\% & 0.94\% & 0.18\% \\
& & Acc (Cons) & 74.55\% & 68.76\% & 57.33\% & 36.39\% & 11.02\% & 2.20\% & 0.75\% & 0.32\% \\
& & Lat (Ours) & 7.76 ± 0.041 & 7.15 ± 0.015 & 7.15 ± 0.007 & 6.36 ± 0.006 & 6.31 ± 0.007 & 5.47 ± 0.007 & 5.72 ± 0.016 & 4.94 ± 0.001 \\
& & Lat (Base) & 8.21 ± 0.013 & 7.87 ± 0.017 & 7.70 ± 0.022 & 6.90 ± 0.008 & 6.96 ± 0.004 & 6.04 ± 0.006 & 6.41 ± 0.012 & 5.53 ± 0.007 \\
& & Lat (Zero) & 7.24 ± 0.010 & 6.81 ± 0.010 & 6.58 ± 0.008 & 5.81 ± 0.008 & 5.86 ± 0.009 & 4.88 ± 0.004 & 5.17 ± 0.021 & 4.32 ± 0.005 \\
\cmidrule{2-11}
& L2 & Acc (Ours) & 75.55\% & 72.75\% & 65.53\% & 50.44\% & 23.15\% & 4.75\% & 1.23\% & 0.24\% \\
& & Acc (Cons) & 75.18\% & 71.49\% & 61.43\% & 42.96\% & 18.67\% & 3.79\% & 1.46\% & 0.28\% \\
& & Lat (Ours) & 7.68 ± 0.007 & 7.23 ± 0.007 & 6.94 ± 0.011 & 6.37 ± 0.010 & 6.39 ± 0.006 & 5.48 ± 0.007 & 5.78 ± 0.004 & 4.87 ± 0.003 \\
& & Lat (Base) & 8.11 ± 0.012 & 7.78 ± 0.012 & 7.62 ± 0.010 & 6.90 ± 0.008 & 6.94 ± 0.009 & 6.05 ± 0.003 & 6.37 ± 0.006 & 5.49 ± 0.006 \\
& & Lat (Zero) & 7.22 ± 0.035 & 6.74 ± 0.013 & 6.56 ± 0.010 & 5.82 ± 0.004 & 5.84 ± 0.002 & 4.87 ± 0.003 & 5.13 ± 0.014 & 4.32 ± 0.006 \\
\cmidrule{2-11}
& LAMP & Acc (Ours) & 66.15\% & 62.55\% & 53.84\% & 43.46\% & 24.45\% & 6.25\% & 1.01\% & 0.18\% \\
& & Acc (Cons) & 64.56\% & 56.08\% & 1.29\% & 0.17\% & 0.11\% & 0.10\% & 0.10\% & 0.10\% \\
& & Lat (Ours) & 6.22 ± 0.006 & 6.47 ± 0.009 & 6.49 ± 0.011 & 6.44 ± 0.005 & 6.55 ± 0.013 & 6.60 ± 0.004 & 6.50 ± 0.007 & 6.92 ± 0.011 \\
& & Lat (Base) & 6.36 ± 0.009 & 6.51 ± 0.012 & 6.59 ± 0.007 & 6.52 ± 0.005 & 6.67 ± 0.009 & 6.70 ± 0.003 & 6.62 ± 0.006 & 7.18 ± 0.013 \\
& & Lat (Zero) & 6.14 ± 0.004 & 6.07 ± 0.005 & 6.54 ± 0.008 & 6.22 ± 0.040 & 6.01 ± 0.017 & 5.58 ± 0.010 & 5.19 ± 0.017 & 4.73 ± 0.014 \\
\midrule
resnet18 
& FPGM & Acc (Ours) & 67.69\% & 54.76\% & 31.62\% & 6.44\% & 2.19\% & 0.57\% & 0.68\% & 0.15\% \\
& & Acc (Cons) & 67.41\% & 55.00\% & 34.71\% & 9.75\% & 4.93\% & 1.74\% & 0.70\% & 0.14\% \\
& & Lat (Ours) & 2.04 ± 0.009 & 2.02 ± 0.004 & 1.90 ± 0.007 & 1.83 ± 0.002 & 1.77 ± 0.005 & 1.69 ± 0.005 & 1.64 ± 0.001 & 1.48 ± 0.002 \\
& & Lat (Base) & 2.13 ± 0.033 & 2.13 ± 0.005 & 2.03 ± 0.004 & 1.95 ± 0.004 & 1.89 ± 0.005 & 1.80 ± 0.003 & 1.75 ± 0.003 & 1.58 ± 0.003 \\
& & Lat (Zero) & 1.99 ± 0.002 & 1.94 ± 0.010 & 1.85 ± 0.009 & 1.77 ± 0.011 & 1.69 ± 0.007 & 1.58 ± 0.010 & 1.53 ± 0.004 & 1.37 ± 0.008 \\
\cmidrule{2-11}
& HRank & Acc (Ours) & 67.02\% & 51.01\% & 24.09\% & 8.82\% & 2.66\% & 0.48\% & 0.34\% & 0.19\% \\
& & Acc (Cons) & 67.01\% & 52.59\% & 29.94\% & 8.51\% & 1.91\% & 0.77\% & 0.54\% & 0.27\% \\
& & Lat (Ours) & 2.03 ± 0.010 & 1.99 ± 0.007 & 1.89 ± 0.008 & 1.83 ± 0.008 & 1.76 ± 0.006 & 1.66 ± 0.004 & 1.62 ± 0.006 & 1.45 ± 0.007 \\
& & Lat (Base) & 2.14 ± 0.010 & 2.16 ± 0.007 & 2.05 ± 0.009 & 1.98 ± 0.008 & 1.91 ± 0.009 & 1.81 ± 0.005 & 1.78 ± 0.007 & 1.59 ± 0.004 \\
& & Lat (Zero) & 2.03 ± 0.010 & 1.97 ± 0.004 & 1.90 ± 0.007 & 1.79 ± 0.006 & 1.72 ± 0.009 & 1.60 ± 0.009 & 1.55 ± 0.003 & 1.40 ± 0.004 \\
\cmidrule{2-11}
& L1 & Acc (Ours) & 67.50\% & 55.76\% & 35.09\% & 15.84\% & 7.33\% & 1.68\% & 0.48\% & 0.23\% \\
& & Acc (Cons) & 67.02\% & 54.53\% & 30.17\% & 13.00\% & 2.91\% & 0.93\% & 0.43\% & 0.27\% \\
& & Lat (Ours) & 2.02 ± 0.002 & 2.01 ± 0.005 & 1.90 ± 0.007 & 1.83 ± 0.005 & 1.77 ± 0.004 & 1.69 ± 0.003 & 1.64 ± 0.004 & 1.47 ± 0.006 \\
& & Lat (Base) & 2.10 ± 0.006 & 2.15 ± 0.024 & 2.03 ± 0.003 & 1.96 ± 0.003 & 1.88 ± 0.001 & 1.81 ± 0.008 & 1.75 ± 0.004 & 1.58 ± 0.007 \\
& & Lat (Zero) & 2.00 ± 0.009 & 1.95 ± 0.012 & 1.85 ± 0.007 & 1.76 ± 0.005 & 1.70 ± 0.005 & 1.59 ± 0.004 & 1.53 ± 0.001 & 1.38 ± 0.006 \\
\cmidrule{2-11}
& L2 & Acc (Ours) & 67.57\% & 54.34\% & 35.54\% & 15.41\% & 4.54\% & 1.28\% & 0.67\% & 0.18\% \\
& & Acc (Cons) & 67.18\% & 55.27\% & 34.29\% & 15.96\% & 5.22\% & 1.61\% & 1.01\% & 0.16\% \\
& & Lat (Ours) & 2.06 ± 0.008 & 2.03 ± 0.006 & 1.91 ± 0.006 & 1.85 ± 0.004 & 1.79 ± 0.009 & 1.71 ± 0.009 & 1.65 ± 0.003 & 1.48 ± 0.008 \\
& & Lat (Base) & 2.12 ± 0.020 & 2.15 ± 0.003 & 2.05 ± 0.025 & 1.96 ± 0.005 & 1.90 ± 0.005 & 1.81 ± 0.009 & 1.77 ± 0.004 & 1.59 ± 0.003 \\
& & Lat (Zero) & 2.01 ± 0.008 & 1.97 ± 0.009 & 1.86 ± 0.004 & 1.79 ± 0.010 & 1.71 ± 0.012 & 1.59 ± 0.005 & 1.55 ± 0.004 & 1.39 ± 0.006 \\
\cmidrule{2-11}
& LAMP & Acc (Ours) & 66.75\% & 56.40\% & 38.07\% & 21.45\% & 7.74\% & 2.71\% & 1.27\% & 0.28\% \\
& & Acc (Cons) & 67.21\% & 58.86\% & 44.88\% & 30.68\% & 17.83\% & 7.71\% & 3.15\% & 0.43\% \\
& & Lat (Ours) & 2.03 ± 0.006 & 1.98 ± 0.004 & 1.61 ± 0.005 & 1.86 ± 0.006 & 1.80 ± 0.007 & 1.74 ± 0.006 & 1.68 ± 0.008 & 1.51 ± 0.004 \\
& & Lat (Base) & 2.06 ± 0.005 & 2.01 ± 0.004 & 1.65 ± 0.002 & 1.90 ± 0.008 & 1.87 ± 0.007 & 1.80 ± 0.003 & 1.74 ± 0.006 & 1.57 ± 0.006 \\
& & Lat (Zero) & 2.02 ± 0.008 & 1.98 ± 0.025 & 1.63 ± 0.006 & 1.87 ± 0.009 & 1.81 ± 0.009 & 1.76 ± 0.008 & 1.67 ± 0.005 & 1.49 ± 0.006 \\
\midrule
resnet50 
& FPGM & Acc (Ours) & 73.93\% & 68.69\% & 52.07\% & 19.21\% & 3.12\% & 0.68\% & 0.59\% & 0.19\% \\
& & Acc (Cons) & 73.84\% & 65.91\% & 45.81\% & 14.15\% & 2.47\% & 0.46\% & 0.58\% & 0.21\% \\
& & Lat (Ours) & 4.92 ± 0.024 & 4.36 ± 0.009 & 4.42 ± 0.015 & 3.60 ± 0.008 & 3.77 ± 0.016 & 3.13 ± 0.007 & 3.55 ± 0.004 & 2.88 ± 0.005 \\
& & Lat (Base) & 5.06 ± 0.013 & 4.72 ± 0.012 & 4.65 ± 0.004 & 3.95 ± 0.007 & 4.15 ± 0.009 & 3.40 ± 0.004 & 3.81 ± 0.013 & 3.14 ± 0.004 \\
& & Lat (Zero) & 4.59 ± 0.014 & 4.20 ± 0.018 & 4.07 ± 0.018 & 3.36 ± 0.000 & 3.54 ± 0.020 & 2.77 ± 0.007 & 3.15 ± 0.005 & 2.50 ± 0.003 \\
\cmidrule{2-11}
& HRank & Acc (Ours) & 73.05\% & 58.48\% & 30.06\% & 6.08\% & 1.08\% & 0.26\% & 0.20\% & 0.10\% \\
& & Acc (Cons) & 72.94\% & 59.27\% & 29.90\% & 3.11\% & 0.76\% & 0.19\% & 0.11\% & 0.10\% \\
& & Lat (Ours) & 4.73 ± 0.007 & 4.40 ± 0.016 & 4.31 ± 0.006 & 4.23 ± 0.009 & 3.87 ± 0.011 & 3.63 ± 0.010 & 3.53 ± 0.009 & 3.09 ± 0.009 \\
& & Lat (Base) & 4.95 ± 0.006 & 4.66 ± 0.006 & 4.59 ± 0.004 & 4.48 ± 0.003 & 4.15 ± 0.004 & 3.94 ± 0.008 & 3.83 ± 0.016 & 3.52 ± 0.011 \\
& & Lat (Zero) & 4.54 ± 0.011 & 4.20 ± 0.011 & 4.04 ± 0.007 & 3.90 ± 0.008 & 3.53 ± 0.010 & 3.30 ± 0.005 & 3.17 ± 0.007 & 2.86 ± 0.005 \\
\cmidrule{2-11}
& L1 & Acc (Ours) & 73.88\% & 68.66\% & 48.79\% & 14.87\% & 2.31\% & 0.83\% & 0.47\% & 0.25\% \\
& & Acc (Cons) & 73.09\% & 65.24\% & 37.39\% & 8.88\% & 1.68\% & 0.40\% & 0.27\% & 0.15\% \\
& & Lat (Ours) & 4.85 ± 0.004 & 4.42 ± 0.015 & 4.37 ± 0.009 & 3.72 ± 0.005 & 3.74 ± 0.009 & 3.03 ± 0.004 & 3.50 ± 0.013 & 2.87 ± 0.003 \\
& & Lat (Base) & 5.04 ± 0.035 & 4.69 ± 0.009 & 4.60 ± 0.012 & 3.96 ± 0.004 & 4.13 ± 0.009 & 3.41 ± 0.007 & 3.76 ± 0.007 & 3.13 ± 0.008 \\
& & Lat (Zero) & 4.57 ± 0.051 & 4.13 ± 0.013 & 4.01 ± 0.011 & 3.34 ± 0.006 & 3.49 ± 0.016 & 2.77 ± 0.002 & 3.10 ± 0.013 & 2.49 ± 0.002 \\
\cmidrule{2-11}
& L2 & Acc (Ours) & 73.87\% & 68.52\% & 50.38\% & 18.51\% & 3.43\% & 0.84\% & 0.52\% & 0.25\% \\
& & Acc (Cons) & 73.90\% & 65.91\% & 41.92\% & 13.03\% & 3.48\% & 0.56\% & 0.51\% & 0.25\% \\
& & Lat (Ours) & 4.87 ± 0.012 & 4.48 ± 0.075 & 4.39 ± 0.004 & 3.74 ± 0.007 & 3.87 ± 0.011 & 3.13 ± 0.001 & 3.51 ± 0.008 & 2.89 ± 0.007 \\
& & Lat (Base) & 5.03 ± 0.011 & 4.69 ± 0.006 & 4.61 ± 0.011 & 3.98 ± 0.004 & 4.12 ± 0.003 & 3.40 ± 0.006 & 3.77 ± 0.002 & 3.14 ± 0.004 \\
& & Lat (Zero) & 4.55 ± 0.017 & 4.19 ± 0.048 & 4.03 ± 0.013 & 3.35 ± 0.005 & 3.50 ± 0.014 & 2.76 ± 0.003 & 3.11 ± 0.009 & 2.49 ± 0.004 \\
\cmidrule{2-11}
& LAMP & Acc (Ours) & 67.83\% & 56.40\% & 49.41\% & 35.34\% & 19.94\% & 12.11\% & 3.70\% & 0.17\% \\
& & Acc (Cons) & 68.32\% & 61.55\% & 44.33\% & 13.28\% & 1.79\% & 0.15\% & 0.20\% & 0.10\% \\
& & Lat (Ours) & 3.64 ± 0.003 & 3.61 ± 0.003 & 3.62 ± 0.002 & 3.58 ± 0.008 & 3.55 ± 0.004 & 3.58 ± 0.005 & 3.55 ± 0.005 & 3.94 ± 0.010 \\
& & Lat (Base) & 3.65 ± 0.005 & 3.64 ± 0.003 & 3.68 ± 0.006 & 3.65 ± 0.004 & 3.65 ± 0.002 & 3.64 ± 0.003 & 3.62 ± 0.005 & 4.08 ± 0.006 \\
& & Lat (Zero) & 3.48 ± 0.007 & 3.43 ± 0.002 & 3.35 ± 0.004 & 3.25 ± 0.006 & 3.79 ± 0.054 & 3.54 ± 0.005 & 3.35 ± 0.007 & 3.38 ± 0.012 \\

        \bottomrule
        \end{tabular}
        \caption{ \textbf{UPSCALE accuracy and latency} across architectures, sparsity levels, and heuristics. Notice that latency for ours is comparable to the ideal, zero-copy reference, much lower than the baseline export's latency. Additionally notice our (unconstrained) accuracy matches or outperforms the baseline constrained accuracy. }
        \label{tab:upscale_vs_naive_latency_2}
    \end{table}\begin{table}
    \centering
    \tiny
    \begin{tabular}{l|l|cccccccccc}
    \toprule
    Model & Heuristic & Stat & 1\% & 5\% & 10\% & 15\% & 20\% & 25\% & 30\% & 40\% \\

\midrule
squeezenet1\_0 
& FPGM & Acc (Ours) & 57.60\% & 42.29\% & 10.70\% & 4.43\% & 1.30\% & 0.33\% & 0.21\% & 0.12\% \\
& & Acc (Cons) & 57.60\% & 30.96\% & 6.22\% & 1.76\% & 0.47\% & 0.19\% & 0.17\% & 0.12\% \\
& & Lat (Ours) & 0.79 ± 0.001 & 0.79 ± 0.002 & 0.78 ± 0.004 & 0.77 ± 0.004 & 0.76 ± 0.002 & 0.74 ± 0.002 & 0.74 ± 0.003 & 0.70 ± 0.003 \\
& & Lat (Base) & 0.86 ± 0.001 & 0.94 ± 0.003 & 0.95 ± 0.003 & 0.94 ± 0.002 & 0.94 ± 0.004 & 0.90 ± 0.002 & 0.91 ± 0.003 & 0.87 ± 0.001 \\
& & Lat (Zero) & 0.79 ± 0.002 & 0.79 ± 0.002 & 0.77 ± 0.004 & 0.76 ± 0.001 & 0.76 ± 0.004 & 0.73 ± 0.002 & 0.72 ± 0.000 & 0.69 ± 0.003 \\
\cmidrule{2-11}
& HRank & Acc (Ours) & 55.99\% & 18.79\% & 1.33\% & 0.36\% & 0.21\% & 0.11\% & 0.10\% & 0.10\% \\
& & Acc (Cons) & 55.99\% & 18.79\% & 1.33\% & 0.36\% & 0.21\% & 0.11\% & 0.10\% & 0.10\% \\
& & Lat (Ours) & 0.78 ± 0.002 & 0.78 ± 0.002 & 0.76 ± 0.003 & 0.75 ± 0.002 & 0.75 ± 0.001 & 0.73 ± 0.004 & 0.72 ± 0.000 & 0.67 ± 0.002 \\
& & Lat (Base) & 0.85 ± 0.003 & 0.88 ± 0.002 & 0.88 ± 0.002 & 0.87 ± 0.003 & 0.87 ± 0.002 & 0.84 ± 0.003 & 0.83 ± 0.002 & 0.78 ± 0.002 \\
& & Lat (Zero) & 0.78 ± 0.003 & 0.78 ± 0.000 & 0.76 ± 0.000 & 0.75 ± 0.001 & 0.76 ± 0.001 & 0.73 ± 0.002 & 0.72 ± 0.002 & 0.67 ± 0.001 \\
\cmidrule{2-11}
& L1 & Acc (Ours) & 55.06\% & 38.67\% & 12.88\% & 2.99\% & 1.06\% & 0.33\% & 0.20\% & 0.09\% \\
& & Acc (Cons) & 55.06\% & 28.22\% & 5.39\% & 1.49\% & 0.67\% & 0.31\% & 0.22\% & 0.11\% \\
& & Lat (Ours) & 0.78 ± 0.001 & 0.78 ± 0.005 & 0.78 ± 0.002 & 0.76 ± 0.001 & 0.76 ± 0.003 & 0.74 ± 0.001 & 0.72 ± 0.003 & 0.70 ± 0.002 \\
& & Lat (Base) & 0.85 ± 0.002 & 0.94 ± 0.003 & 0.94 ± 0.002 & 0.94 ± 0.003 & 0.93 ± 0.002 & 0.90 ± 0.003 & 0.89 ± 0.002 & 0.86 ± 0.002 \\
& & Lat (Zero) & 0.78 ± 0.001 & 0.79 ± 0.003 & 0.77 ± 0.003 & 0.76 ± 0.003 & 0.76 ± 0.004 & 0.73 ± 0.001 & 0.71 ± 0.001 & 0.68 ± 0.001 \\
\cmidrule{2-11}
& L2 & Acc (Ours) & 57.61\% & 44.45\% & 11.36\% & 3.02\% & 1.16\% & 0.30\% & 0.15\% & 0.12\% \\
& & Acc (Cons) & 57.61\% & 31.25\% & 5.47\% & 1.23\% & 0.26\% & 0.24\% & 0.13\% & 0.19\% \\
& & Lat (Ours) & 0.79 ± 0.003 & 0.78 ± 0.001 & 0.77 ± 0.002 & 0.75 ± 0.004 & 0.76 ± 0.003 & 0.73 ± 0.001 & 0.73 ± 0.003 & 0.70 ± 0.002 \\
& & Lat (Base) & 0.85 ± 0.002 & 0.93 ± 0.003 & 0.94 ± 0.002 & 0.93 ± 0.001 & 0.93 ± 0.002 & 0.90 ± 0.002 & 0.90 ± 0.003 & 0.86 ± 0.003 \\
& & Lat (Zero) & 0.79 ± 0.004 & 0.78 ± 0.001 & 0.77 ± 0.003 & 0.75 ± 0.003 & 0.75 ± 0.001 & 0.72 ± 0.003 & 0.72 ± 0.002 & 0.68 ± 0.002 \\
\cmidrule{2-11}
& LAMP & Acc (Ours) & 56.12\% & 29.19\% & 13.54\% & 3.87\% & 0.95\% & 0.47\% & 0.19\% & 0.15\% \\
& & Acc (Cons) & 56.39\% & 41.67\% & 26.35\% & 12.60\% & 4.76\% & 1.97\% & 0.95\% & 0.26\% \\
& & Lat (Ours) & 0.77 ± 0.001 & 0.77 ± 0.003 & 0.76 ± 0.003 & 0.75 ± 0.000 & 0.74 ± 0.004 & 0.74 ± 0.002 & 0.72 ± 0.001 & 0.73 ± 0.001 \\
& & Lat (Base) & 0.83 ± 0.002 & 0.83 ± 0.003 & 0.83 ± 0.002 & 0.82 ± 0.002 & 0.82 ± 0.003 & 0.82 ± 0.002 & 0.79 ± 0.002 & 0.80 ± 0.003 \\
& & Lat (Zero) & 0.77 ± 0.004 & 0.77 ± 0.002 & 0.77 ± 0.003 & 0.77 ± 0.002 & 0.75 ± 0.000 & 0.75 ± 0.002 & 0.73 ± 0.001 & 0.74 ± 0.001 \\
\midrule
squeezenet1\_1 
& FPGM & Acc (Ours) & 57.80\% & 41.09\% & 8.25\% & 1.04\% & 0.58\% & 0.23\% & 0.19\% & 0.16\% \\
& & Acc (Cons) & 57.80\% & 28.85\% & 4.42\% & 0.78\% & 0.27\% & 0.16\% & 0.13\% & 0.10\% \\
& & Lat (Ours) & 0.66 ± 0.002 & 0.67 ± 0.002 & 0.66 ± 0.001 & 0.67 ± 0.003 & 0.66 ± 0.003 & 0.65 ± 0.003 & 0.65 ± 0.002 & 0.62 ± 0.002 \\
& & Lat (Base) & 0.72 ± 0.003 & 0.80 ± 0.006 & 0.83 ± 0.002 & 0.83 ± 0.003 & 0.83 ± 0.002 & 0.81 ± 0.002 & 0.81 ± 0.001 & 0.78 ± 0.003 \\
& & Lat (Zero) & 0.66 ± 0.002 & 0.67 ± 0.002 & 0.66 ± 0.004 & 0.66 ± 0.003 & 0.67 ± 0.001 & 0.66 ± 0.004 & 0.66 ± 0.002 & 0.62 ± 0.001 \\
\cmidrule{2-11}
& HRank & Acc (Ours) & 56.09\% & 20.69\% & 1.06\% & 0.28\% & 0.14\% & 0.14\% & 0.13\% & 0.10\% \\
& & Acc (Cons) & 56.09\% & 20.69\% & 1.06\% & 0.28\% & 0.14\% & 0.14\% & 0.13\% & 0.10\% \\
& & Lat (Ours) & 0.67 ± 0.002 & 0.67 ± 0.004 & 0.67 ± 0.002 & 0.65 ± 0.002 & 0.67 ± 0.001 & 0.66 ± 0.004 & 0.67 ± 0.002 & 0.62 ± 0.001 \\
& & Lat (Base) & 0.74 ± 0.001 & 0.78 ± 0.004 & 0.79 ± 0.002 & 0.76 ± 0.002 & 0.79 ± 0.003 & 0.77 ± 0.003 & 0.77 ± 0.003 & 0.73 ± 0.004 \\
& & Lat (Zero) & 0.67 ± 0.002 & 0.67 ± 0.002 & 0.67 ± 0.002 & 0.65 ± 0.001 & 0.67 ± 0.001 & 0.65 ± 0.003 & 0.66 ± 0.002 & 0.62 ± 0.003 \\
\cmidrule{2-11}
& L1 & Acc (Ours) & 57.83\% & 35.67\% & 12.30\% & 1.41\% & 0.70\% & 0.16\% & 0.12\% & 0.13\% \\
& & Acc (Cons) & 57.83\% & 21.78\% & 7.73\% & 0.98\% & 0.39\% & 0.21\% & 0.16\% & 0.12\% \\
& & Lat (Ours) & 0.66 ± 0.001 & 0.67 ± 0.002 & 0.67 ± 0.002 & 0.67 ± 0.004 & 0.66 ± 0.002 & 0.64 ± 0.001 & 0.63 ± 0.002 & 0.61 ± 0.003 \\
& & Lat (Base) & 0.72 ± 0.004 & 0.81 ± 0.001 & 0.84 ± 0.005 & 0.84 ± 0.002 & 0.83 ± 0.002 & 0.81 ± 0.002 & 0.82 ± 0.003 & 0.77 ± 0.002 \\
& & Lat (Zero) & 0.66 ± 0.002 & 0.68 ± 0.003 & 0.67 ± 0.002 & 0.66 ± 0.001 & 0.67 ± 0.003 & 0.66 ± 0.001 & 0.65 ± 0.003 & 0.62 ± 0.006 \\
\cmidrule{2-11}
& L2 & Acc (Ours) & 57.71\% & 39.84\% & 9.44\% & 0.88\% & 0.35\% & 0.19\% & 0.13\% & 0.12\% \\
& & Acc (Cons) & 57.71\% & 29.70\% & 5.80\% & 1.41\% & 0.32\% & 0.12\% & 0.14\% & 0.09\% \\
& & Lat (Ours) & 0.67 ± 0.003 & 0.67 ± 0.001 & 0.67 ± 0.002 & 0.67 ± 0.002 & 0.66 ± 0.002 & 0.65 ± 0.003 & 0.65 ± 0.002 & 0.62 ± 0.003 \\
& & Lat (Base) & 0.73 ± 0.003 & 0.81 ± 0.001 & 0.84 ± 0.002 & 0.85 ± 0.002 & 0.82 ± 0.001 & 0.82 ± 0.003 & 0.81 ± 0.003 & 0.78 ± 0.002 \\
& & Lat (Zero) & 0.67 ± 0.002 & 0.68 ± 0.001 & 0.67 ± 0.003 & 0.67 ± 0.001 & 0.67 ± 0.004 & 0.66 ± 0.001 & 0.65 ± 0.001 & 0.61 ± 0.002 \\
\cmidrule{2-11}
& LAMP & Acc (Ours) & 53.77\% & 44.22\% & 25.55\% & 8.22\% & 1.76\% & 0.71\% & 0.26\% & 0.11\% \\
& & Acc (Cons) & 53.57\% & 45.79\% & 29.27\% & 11.53\% & 2.76\% & 1.22\% & 0.45\% & 0.15\% \\
& & Lat (Ours) & 0.67 ± 0.001 & 0.66 ± 0.002 & 0.66 ± 0.003 & 0.66 ± 0.002 & 0.67 ± 0.001 & 0.65 ± 0.003 & 0.69 ± 0.002 & 0.69 ± 0.003 \\
& & Lat (Base) & 0.72 ± 0.001 & 0.72 ± 0.002 & 0.72 ± 0.001 & 0.72 ± 0.004 & 0.73 ± 0.002 & 0.71 ± 0.001 & 0.75 ± 0.003 & 0.75 ± 0.003 \\
& & Lat (Zero) & 0.67 ± 0.003 & 0.67 ± 0.002 & 0.66 ± 0.003 & 0.66 ± 0.001 & 0.65 ± 0.000 & 0.67 ± 0.004 & 0.65 ± 0.001 & 0.70 ± 0.002 \\

        \bottomrule
        \end{tabular}
        \caption{ \textbf{UPSCALE accuracy and latency} across architectures, sparsity levels, and heuristics. Notice that latency for ours is comparable to the ideal, zero-copy reference, much lower than the baseline export's latency. Additionally notice our (unconstrained) accuracy matches or outperforms the baseline constrained accuracy. }
        \label{tab:upscale_vs_naive_latency_3}
    \end{table}

\end{document}